\documentclass{article} % For LaTeX2e
\usepackage{iclr2026_conference,times}

\iclrfinalcopy % 1. 开启完稿模式：关闭匿名（显示作者），去掉左侧 review 行数

\renewcommand{\headrulewidth}{0pt} % 清除页眉下方的横线（如有）
\usepackage{amsmath,amsfonts,bm}

\def\eqref#1{equation~\ref{#1}}
\def\1{\bm{1}}

\DeclareMathAlphabet{\mathsfit}{\encodingdefault}{\sfdefault}{m}{sl}
\SetMathAlphabet{\mathsfit}{bold}{\encodingdefault}{\sfdefault}{bx}{n}

\usepackage{graphicx}
\usepackage{subcaption}
\usepackage{booktabs}
\usepackage{multirow}
\usepackage{amsmath,amssymb,amsthm,mathtools}
\usepackage{hyperref}
\usepackage{url}
\usepackage{enumitem}
\usepackage{xcolor}
\usepackage{array}
\usepackage{makecell}
\usepackage{adjustbox}
\usepackage{tabularx}
\usepackage{siunitx}
\usepackage{microtype}
\usepackage{cleveref}

\graphicspath{{figures/}{./figures/}}
\newcommand{\tabfit}[1]{\begin{adjustbox}{max width=\linewidth}#1\end{adjustbox}}
\newcommand{\xii}{\xi}
\newcommand{\deltae}{\delta}
\newcommand{\Ptilde}{\widetilde P}
\newcommand{\Lc}{L^{(\gamma)}}
\newcommand{\taucommit}{\tau_{\mathrm{commit}}^{(\gamma)}}
\newcommand{\tauonset}{\tau_{\mathrm{onset}}}
\newcommand{\Aset}{\mathcal A}

\newcommand{\yes}{\mathrm{yes}}
\newcommand{\no}{\mathrm{no}}

\title{When Does a Language Model Commit?\\
A Finite-Answer Theory of Pre-Verbalization Commitment}

\author{%
  Long Zhang\thanks{Corresponding authors: longzhang@scut.edu.cn, cschenwn@scut.edu.cn} \quad 
  Wei-neng Chen\footnotemark[1]\quad 
  Feng-feng Wei\quad 
  Zi-bo Qin \\
  School of Computer Science and Engineering \\
  South China University of Technology \\
  Guangzhou City, Guangdong Province, China \\
  \texttt{\{longzhang, cschenwn\}@scut.edu.cn} \\
}

\begin{document}
\maketitle

\begin{abstract}
Language models often generate reasoning before giving a final answer, but the visible answer does not reveal when the model's answer preference became stable. We study this question through a narrow computable object: \emph{finite-answer preference stabilization}. For a model state and specified answer verbalizers, we project the model's own continuation probabilities onto a finite answer set; in binary tasks this yields an exact log-odds code, $\delta(\xi)=S_\theta(\mathrm{yes}\mid\xi)-S_\theta(\mathrm{no}\mid\xi)$. This target defines parser-based answer onset, retrospective stabilization time, and lead without relying on greedy rollouts or learned probes. In controlled delayed-verdict tasks with Qwen3-4B-Instruct, the contextual finite-answer projection stabilizes before the answer is parseable, with 17--31 token mean lead in the main templates and positive, shorter lead in a parser-clean replication. The signal tracks the model's eventual output rather than truth, is linearly recoverable from compact hidden summaries, is partly separable from cursor progress, and transfers as shared information without a single invariant coordinate. Diagnostics separate the measurement from online stopping, verbalizer-free belief, and causal answer control; exact steering shows local sensitivity of $\delta$ but not reliable generation control.
\end{abstract}

\vspace{0.5em}
\noindent\textbf{Keywords:} Large Language Models, Latent Reasoning, Chain-of-Thought, Mechanistic Interpretability, Pre-verbalization Commitment
\vspace{1em}

\section{Introduction}

Language models often write intermediate reasoning before final answers, and such traces can improve performance \citep{wei2022chain,kojima2022large,wang2023selfconsistency}. But the visible chronology of a response need not match the chronology of answer preference. If a model continues writing after its answer preference has stabilized, then some visible reasoning occurs in a post-commitment regime, a possibility closely related to concerns about rationale faithfulness and latent reasoning \citep{turpin2023language,lanham2023measuring,reasoningtheater2026,multilinguallatent2026}.

The difficulty is that commitment is not a token. The final verdict tells us what the model eventually said, not when that answer became preferred. Greedy rollouts are tautological on deterministic trajectories, and hidden-state probes are incomplete: probes can reveal accessible information without defining the model's current preference or causal computation \citep{alain2017understanding,hewitt2019designing,belinkov2022probing}. These approaches conflate the ground-truth answer, the answer eventually generated, and the model's state-dependent probability mass over possible answer continuations.

We define commitment by separating these objects. Given an autoregressive state, the model itself assigns probabilities to all finite continuations. Given a finite answer set and answer verbalizers, we project those continuation probabilities onto the answer set, following the general idea that language-model probabilities can be mapped to finite labels through specified verbalizers \citep{schick2021exploiting,gao2021making,zhao2021calibrate}. In binary tasks, this projection yields an exact scalar log-odds, $\delta(\xi)=S_\theta(\yes\mid\xi)-S_\theta(\no\mid\xi)$, which determines the induced answer distribution, winner, margin, and entropy. This scalar is not a probe target chosen after the fact; it is a computable function of the model's own conditional distribution.

This lets us ask a temporal question. During a delayed-verdict generation, we track $\delta_t$ before any answer is externally visible. We define answer onset as the first token at which a parser can determine the answer, commitment time as the first pre-onset position where the induced winner matches the eventual answer with sufficient margin and does not later flip, and lead as the number of tokens between commitment and onset. A positive lead means that the model's finite-answer projection has committed before the answer is verbalized.

We study this framework in controlled delayed-verdict tasks. These tasks are deliberately simple: they make onset detection reliable, allow exact continuation scoring, and separate answer commitment from template progress. This simplicity is a feature rather than a limitation of the measurement problem. The goal is not to simulate open-ended deliberation in full generality, but to build a clean clock for one narrow object: when the model's own finite-answer continuation preference becomes stable.

This framing also fixes the scope of the paper. We do not claim to measure unrestricted semantic belief, to prove that written reasoning is unfaithful, or to identify a complete causal circuit for decision formation. Instead, we ask a smaller question that can be answered exactly: given specified answer verbalizers, does the model's induced finite-answer distribution stabilize before any answer is parseable, and how is that state-level quantity represented?

\begin{figure}[!htbp]
\centering
\includegraphics[width=0.90\textwidth, height=0.85\textheight, keepaspectratio]{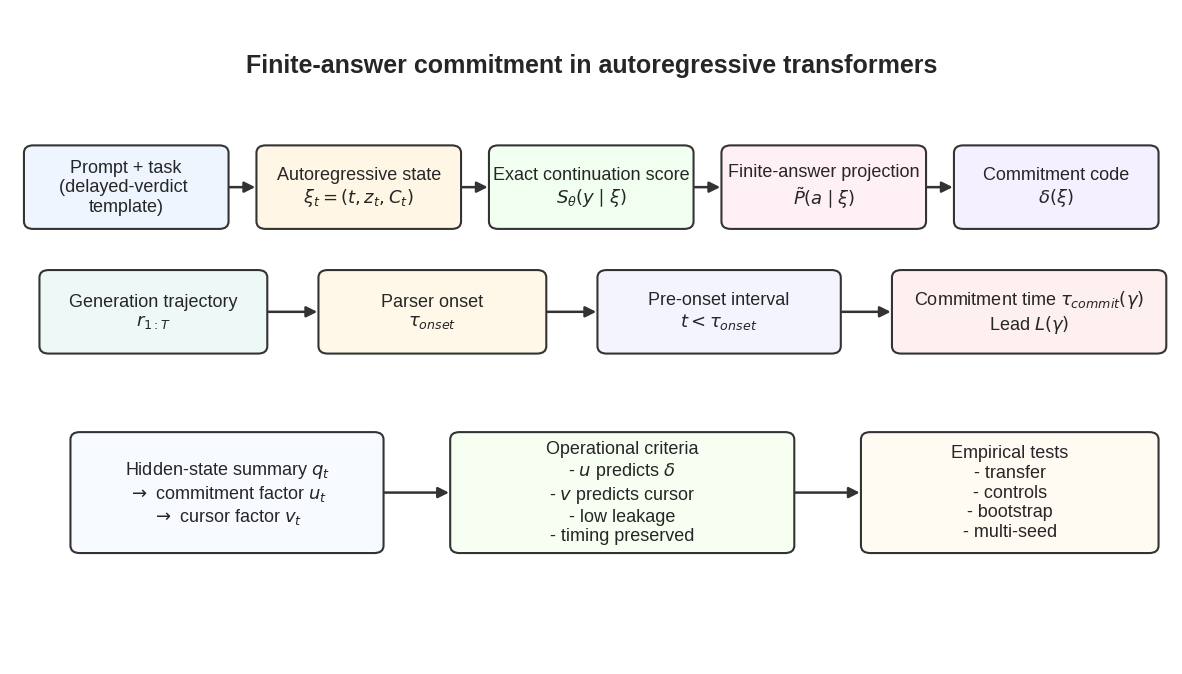}
\caption{\textbf{Finite-answer commitment as a state-level measurement.}
Given an autoregressive state, we score finite answer verbalizers using the model's own continuation probabilities, project these scores onto a finite answer set, and obtain a binary commitment code $\delta(\xi)$. Along a delayed-verdict trajectory, a parser defines answer onset, while the commitment code defines commitment time and lead. Hidden-state probes and factorization experiments are downstream tests of whether this exact target is recoverable and separable from realization progress.}
\label{fig:framework}
\end{figure}

\Cref{fig:framework} summarizes the measurement pipeline. The central methodological point is that commitment is defined before probing: the probe tests whether the exact finite-answer target is recoverable from a compact state summary, but it does not define the target.

Our results show a consistent picture. The surprising fact is not merely that the final answer is predictable from hidden states; it is that the model's own finite-answer continuation distribution can become stable before any answer token is externally parseable. Qwen3-4B's contextual finite-answer projection stabilizes on the eventual answer across prompt, verbalizer, and task-family shifts. The exact commitment code is linearly recoverable from compact hidden-state summaries. However, cross-condition transfer reveals shared information without a single zero-shot invariant linear coordinate, echoing broader evidence that linearly accessible model information can be geometrically structured yet prompt- or distribution-dependent \citep{marks2024geometry}. Commitment can also be operationally separated from realization progress through low-dimensional factors validated by controls, transfer, bootstrap tests, and multi-seed evaluation. Finally, quality-gated and causal-sensitivity diagnostics clarify the scope of the claim: parser-clean templates preserve positive lead but shorten it, calibrated online rules can predict final answers in held-out trajectories without being identical to the retrospective definition, and exact activation steering locally shifts $\delta$ without reliably flipping generated answers \citep{meng2022locating,geiger2021causal}.

\section{Related Work}

Chain-of-thought prompting shows that language models often benefit from writing intermediate reasoning before final answers \citep{wei2022chain,kojima2022large,wang2023selfconsistency}. A central question is whether such rationales faithfully reflect the computation that determines the answer \citep{turpin2023language,lanham2023measuring}. Recent work further suggests that models can exhibit latent or performative reasoning patterns, where internal answer-related information and visible reasoning traces need not evolve in lockstep \citep{reasoningtheater2026,multilinguallatent2026}. Our question is complementary and narrower. We do not directly judge whether a rationale is faithful; instead, we measure when the model's finite-answer continuation preference stabilizes relative to the visible answer.

Linear probes and representation analyses test whether internal states encode task variables \citep{alain2017understanding,belinkov2022probing,hewitt2019designing,hewitt2021conditional,ravfogel2020null,elazar2021amnesic}. Work on latent knowledge and self-knowledge further distinguishes stated answers from internally accessible information \citep{burns2023discovering,azaria2023internal,li2023inference,kadavath2022language,marks2024geometry}. Related self-reflection work asks whether models can verbalize or summarize their own internal answer distributions \citep{selfreflect2026}. These works motivate the possibility that answer-related information exists before it is verbalized. Our distinction is that the target is not supplied by an external label, self-report, or learned probe: $\delta(\xi)$ is an exact continuation-score projection induced by the model's own conditional distribution.

Work on in-advance correctness directions asks whether pre-answer activations predict whether the model will be correct. Our target is different in three ways. First, $\delta(\xi)$ is model-defined rather than label-defined: it is computed from continuation probabilities over answer verbalizers. Second, it tracks the model's eventual answer rather than ground truth, which is why wrong answers can also be retrospectively committed. Third, we measure a trajectory-level time of stabilization relative to parser-defined answer onset, rather than only asking whether a pre-answer activation contains a supervised correctness signal. These questions are complementary: correctness probes ask whether the model will be right; finite-answer commitment asks when the model's own answer preference becomes stable.

Logit-lens and tuned-lens methods reveal how token predictions evolve across layers \citep{belrose2023eliciting,logitlens4llms2025}. Our approach asks an analogous but temporally different question. Rather than reading a raw next-token distribution at a single position or layer, we track a finite-answer projection across autoregressive time and compare its stabilization point to parser-defined answer onset. In short, lens methods ask how token predictions evolve inside a forward pass; our measurement asks when the model's answer-level continuation distribution stabilizes along a generated trajectory. This turns evolving prediction into a trajectory-level measurement of commitment time and lead.

Prompt-based classification maps language-model probabilities over verbalizers to finite labels \citep{schick2021exploiting,gao2021making}, and calibration work shows that such mappings can depend on prompt and verbalizer choice \citep{zhao2021calibrate}. We use verbalizers in a temporal role. At each pre-onset autoregressive state, we marginalize over answer verbalizers to obtain a finite-answer distribution. This makes the measured object explicit and also makes its limitation explicit: it is a verbalizer-conditioned projection, not a verbalizer-free semantic belief.

Mechanistic interpretability and causal intervention studies examine how activations mediate model behavior \citep{elhage2021mathematical,olsson2022context,wang2023interpretability,meng2022locating,geiger2021causal}. Our main contribution is upstream of a full circuit account: we define a precise state-level target and measure its temporal stabilization. We include a small exact-scoring intervention pilot to test local causal sensitivity of this target, but we do not claim causal control over final-answer generation.

\section{Finite-Answer Commitment}
\subsection{Continuation scores and finite-answer projection}

Let $\theta$ be a frozen decoder-only language model. Given a prefix-induced autoregressive state $\xii$ and a finite continuation $y=(y_1,\ldots,y_T)$, define the exact continuation score
\[
S_\theta(y\mid\xii)
=
\sum_{t=1}^T
\log p_\theta(y_t\mid \xii,y_{<t}).
\]
When $\xii$ is induced by prefix $x_{1:n}$, this equals $\log p_\theta(y\mid x_{1:n})$. Appendix~\ref{app:math} expands this notation in terms of state updates and caches.

Let $\Aset=\{a_1,\ldots,a_K\}$ be a finite answer set, and let $\Sigma(a)$ be a finite set of verbalizers for answer $a$, as in verbalizer-based language-model classification \citep{schick2021exploiting,gao2021making}. We define the answer-level score by marginalizing over verbalizers:
\[
S_\theta(a\mid\xii)
=
\log\sum_{y\in\Sigma(a)}\exp S_\theta(y\mid\xii).
\]
The state-induced finite-answer distribution is
\[
\Ptilde_{\theta,\Aset}(a_i\mid\xii)
=
\frac{\exp S_\theta(a_i\mid\xii)}
{\sum_j\exp S_\theta(a_j\mid\xii)}.
\]
This is not an unrestricted semantic belief distribution. It is the model's probability mass projected onto a specified finite answer set through specified verbalizers.

We intentionally score complete verbalizer continuations by their autoregressive log-probability, rather than by per-token average log-probability. This choice follows the definition: the object is probability mass assigned to finite answer continuations, and length-normalizing would define a different projection. At the same time, this means the measured quantity is verbalizer-conditioned and can depend on the length and structure of $\Sigma(a)$. We mitigate this by using condition-matched contextual verbalizers, by including prompt and verbalizer shifts, and by reporting bare-label versus contextual diagnostics. We therefore interpret $\delta(\xi)$ as an exact projection for specified verbalizers, not as a verbalizer-free semantic belief.

\subsection{Binary commitment code}

For binary tasks with $\Aset=\{\yes,\no\}$, define
\[
\deltae(\xii)=S_\theta(\yes\mid\xii)-S_\theta(\no\mid\xii).
\]
Then $\Ptilde(\yes\mid\xii)=\sigma(\deltae(\xii))$ and $\Ptilde(\no\mid\xii)=1-\sigma(\deltae(\xii))$. Thus $\deltae(\xii)$ determines the induced answer distribution, winner, margin, and entropy. It is an exact scalar summary of finite-answer commitment.

\subsection{Answer onset, commitment time, and lead}

Let the model generate response tokens $r_{1:T}$. Let $\pi$ be a parser that maps response prefixes to either an answer in $\Aset$ or undefined. The answer onset is
\[
\tauonset=\min\{t:\pi(r_{1:t})\text{ is defined}\}.
\]
For $t<\tauonset$, we compute $\delta_t=\deltae(\xii_t)$ along the model's actual generation trajectory, and let $a_t$ denote the induced finite-answer winner at state $t$.

Let $a^\star$ be the final parsed answer. For threshold $\gamma>0$, the commitment time is
\[
\taucommit
=
\min
\left\{
t<\tauonset:
a_t=a^\star,\;
|\delta_t|\ge\gamma,\;
a_s=a^\star\;\forall s\in[t,\tauonset)
\right\}.
\]
The lead is
\[
\Lc=\tauonset-\taucommit.
\]
Positive lead means that the model's finite-answer projection has stabilized on the eventual answer before the answer is externally visible. In the main experiments we use $\gamma=2$, corresponding to projected binary confidence $\sigma(2)\approx0.881$.

\subsection{Why greedy rollouts are tautological}

A deterministic greedy rollout from an intermediate state on the same greedy trajectory reproduces the suffix already generated. Therefore a constant greedy final answer along a greedy path does not show early commitment; it is a consequence of determinism. Our commitment code avoids this tautology by measuring the probability-weighted finite-answer projection from each pre-onset state. Appendix~\ref{app:math} gives the full proof.

\section{Delayed-Verdict Evaluation}

\subsection{Tasks and conditions}

We evaluate the framework on controlled delayed-verdict tasks using Qwen3-4B-Instruct-2507 \citep{qwen2025qwen3}. Unless otherwise stated, trajectories are generated by deterministic greedy decoding. This choice makes the generated path reproducible and makes the greedy-rollout tautology explicit; stochastic decoding is an important extension but is not part of the main evaluation. The main conditions are designed to vary prompt form, answer verbalizer, and task family while preserving reliable answer-onset annotation.

\begin{table}[t]
\centering
\caption{\textbf{Main delayed-verdict conditions.}
The main Qwen3 experiment uses four controlled condition families. Each condition yields pre-onset states from which we compute exact finite-answer commitment scores.}
\label{tab:conditions}
\tabfit{
\begin{tabular}{lrrl}
\toprule
Condition & Samples & Pre-onset states & Variation \\
\midrule
Canonical & 96 & 4,137 & parity template \\
Prompt shift & 96 & 5,161 & prompt wording and line labels \\
Verbalizer shift & 96 & 3,960 & answer words \\
Task-family shift & 64 & 2,339 & arithmetic comparison \\
\bottomrule
\end{tabular}
}
\end{table}

The canonical task asks whether $a+b+c+d$ is even and requires a five-line response ending with \texttt{Verdict: yes} or \texttt{Verdict: no}. Prompt shift changes wording and line labels while preserving the parity task. Verbalizer shift replaces \texttt{yes/no} with \texttt{affirmative/negative}. Task-family shift changes the task to whether $(a+b)>(c+d)$. Appendix~\ref{app:expdetails} lists all prompts, parsers, and verbalizers.

\subsection{Exact scoring and hidden summaries}

For every pre-onset state, we compute $S_\theta(\yes\mid\xii_t)$ and $S_\theta(\no\mid\xii_t)$ using teacher-forced continuation scoring over condition-specific verbalizer sets. We then compute $\delta_t$. We extract two compact summaries:
\begin{itemize}
\item \texttt{last\_L21}: the last-position hidden state at layer 21.
\item \texttt{concat\_selected}: the concatenation of selected last-position layer states.
\end{itemize}
Layer 21 is used as a compact mid-to-late anchor; Appendix~\ref{app:layer} shows that the result is not layer cherry-picking.

The experiments ask whether finite-answer commitment precedes answer onset, whether the exact code is recoverable from hidden summaries, whether linear coordinates transfer across conditions, and whether commitment can be separated from surface realization progress. Appendix~\ref{app:sanity} reports implementation sanity checks.

\section{Pre-Verbalization Commitment}

The first empirical question is whether the finite-answer projection stabilizes before any answer is visible. This is stronger than asking whether the eventual answer is predictable from hidden states: the target is the model's own continuation distribution, and the time reference is the parser-defined onset of the answer. A single trajectory makes the measurement intuitive: the final verdict appears only at answer onset, while the finite-answer commitment code can become high-margin and remain aligned with the eventual answer earlier. Representative and distributional trajectory views are in Appendix~\ref{app:trajectory_anatomy}.

\subsection{Aggregate lead across conditions}

\begin{figure}[t]
\centering
\includegraphics[width=\linewidth]{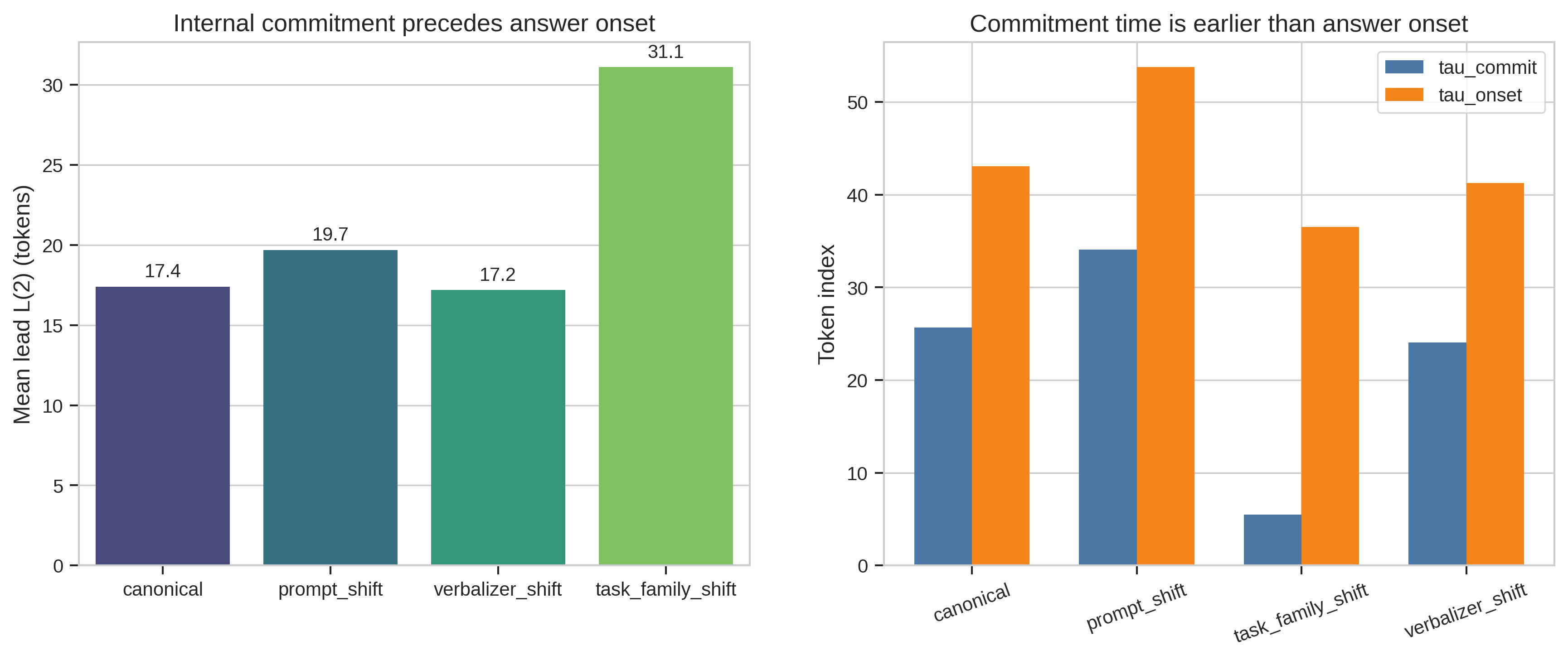}
\caption{\textbf{Finite-answer preference stabilization precedes answer onset across main conditions.}
Left: mean lead at $\gamma=2$ is positive in every condition. Right: mean stabilization time occurs before mean answer onset, showing that the model's finite-answer projection stabilizes before the answer becomes externally visible. The task-family shift has lower task accuracy but still high winner-matches-final, indicating stabilization on the model's eventual answer rather than on ground truth.}
\label{fig:mainphenom}
\end{figure}

\Cref{tab:phenomenon} and \Cref{fig:mainphenom} report the aggregate phenomenon. Across all main conditions, retrospective stabilization occurs before answer onset, with grouped-bootstrap intervals bounded away from zero. Thus the effect is not a single illustrative trajectory, but a trajectory-level regularity.

\begin{table}[t]
\centering
\caption{\textbf{Pre-verbalization finite-answer preference stabilization in the main Qwen3 conditions.}
At threshold $\gamma=2$, retrospective stabilization occurs before parsed answer onset in every main condition. Confidence intervals are grouped bootstrap 95\% intervals over trajectories, not token states. Lead is measured in generated tokens. Winner-matches-final refers to agreement between the final pre-onset finite-answer winner and the model's eventual parsed answer, not necessarily ground truth.}
\label{tab:phenomenon}
\tabfit{
\begin{tabular}{lrrrrrr}
\toprule
Condition & Samples & Acc. & Winner=final & Commit rate & Mean onset & Mean lead \\
\midrule
Canonical & 96 & 1.000 & 1.000 & 1.000 & 43.09 & 17.41 [14.90, 20.19] \\
Prompt shift & 96 & 1.000 & 1.000 & 1.000 & 53.76 & 19.69 [16.61, 22.83] \\
Verbalizer shift & 96 & 1.000 & 1.000 & 1.000 & 41.25 & 17.19 [14.58, 20.05] \\
Task-family shift & 64 & 0.844 & 1.000 & 0.969 & 36.55 & 31.10 [28.33, 33.60] \\
\bottomrule
\end{tabular}
}
\end{table}

The distinction between eventual answer and ground truth is important. In task-family shift, sample accuracy is 0.844, yet winner-matches-final is 1.000 among committed trajectories. The commitment code tracks the model's own eventual answer, including cases where that answer is wrong.

\subsection{The signal tracks eventual output, not truth}

Finite-answer stabilization is not a truth detector. It measures the model's projected preference over specified continuations, and can stabilize on an incorrect eventual answer. In task-family shift, 8 wrong trajectories are retrospectively committed to the model's eventual wrong answer, with mean lead 18.63 tokens; full breakdowns are in Appendix~\ref{app:additional_diagnostics}. The main threshold is also not fragile: Appendix~\ref{app:threshold} shows that increasing $\gamma$ delays commitment but preserves positive lead in all conditions.

\section{The Commitment Code Is Recoverable, But Coordinates Shift}

\subsection{Within-condition and pooled recovery}

We next separate two questions that are often conflated: whether commitment information is present, and whether it occupies the same linear coordinate across conditions. Ridge readouts trained within each condition recover $\delta_t$ with high fidelity, showing that the exact code is accessible from compact hidden summaries. Pooled readouts across canonical, prompt-shift, and verbalizer-shift conditions also perform well, showing that the information is shared enough to support a common supervised readout.

\begin{figure}[t]
\centering
\includegraphics[width=\linewidth]{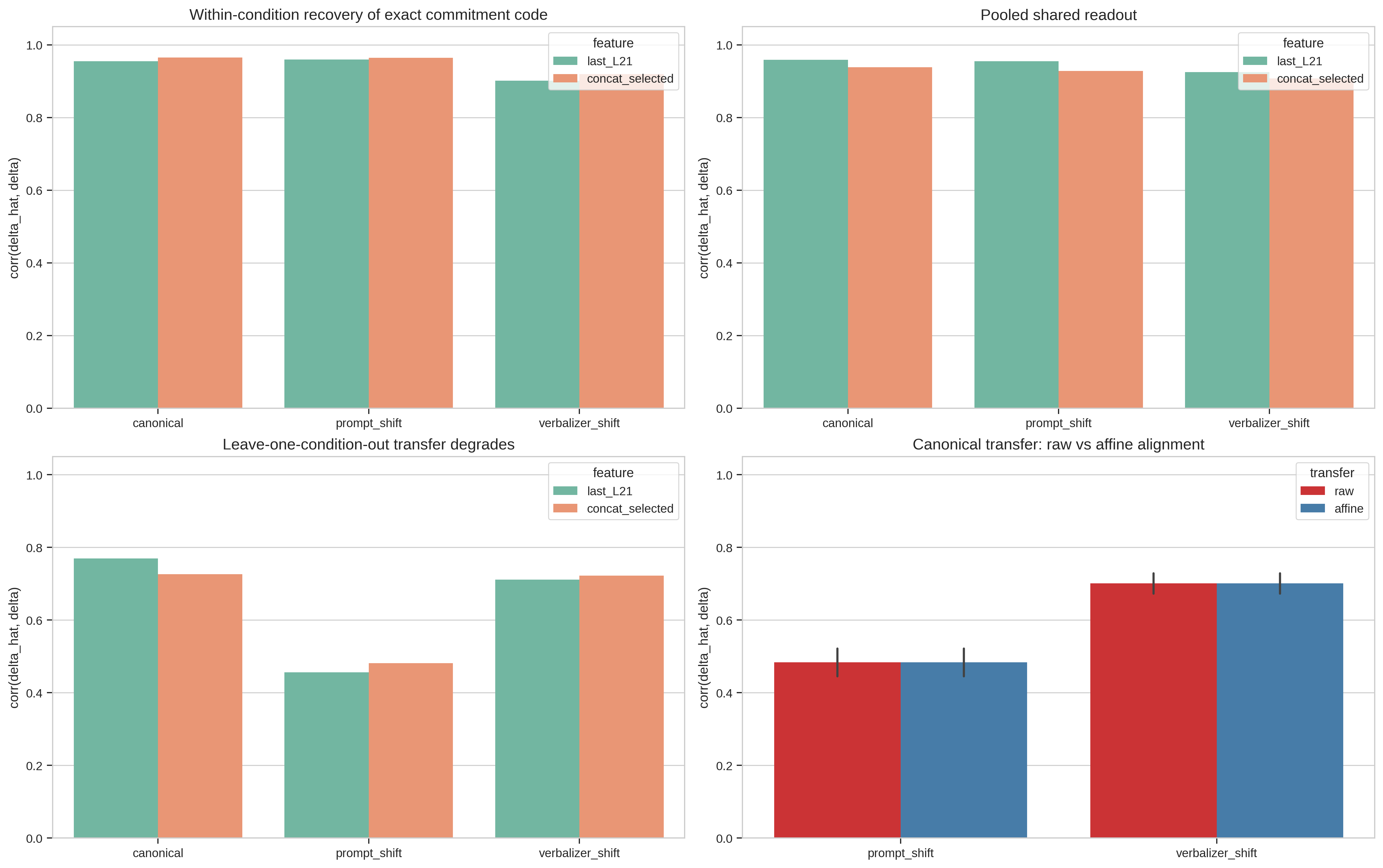}
\caption{\textbf{Commitment is recoverable, but not as a universal zero-shot direction.}
Within-condition and pooled readouts recover the exact commitment code with high fidelity, showing that commitment information is accessible from compact hidden summaries. Leave-one-condition-out and canonical-transfer readouts degrade, especially under prompt shift. Thus commitment information is shared across conditions, but its linear coordinates are condition-dependent.}
\label{fig:lowdim}
\end{figure}

\Cref{fig:lowdim} summarizes the key pattern. The top row shows that both within-condition and pooled readouts strongly recover the exact code. This means commitment information is present in compact summaries and is shared across neighboring conditions.

Representative within-condition correlations are 0.955 for canonical, 0.950 for prompt shift, and 0.929 for verbalizer shift using \texttt{last\_L21}; \texttt{concat\_selected} is similar or slightly better. Appendix~\ref{app:fulltables} provides full tables.

\subsection{Transfer failures distinguish information from coordinates}

The contrast between pooled and leave-one-condition-out readouts is the key diagnostic. The logic is by exclusion. If commitment information were absent under shifts, pooled training would fail. If a single invariant linear coordinate carried the code, leave-one-condition-out transfer would succeed. We observe neither extreme: pooled training succeeds, but zero-shot transfer degrades. Therefore the most consistent interpretation is that related conditions share finite-answer information while embedding it in condition-dependent coordinates.

This distinction is central. It explains why a direction found in one prompt format can be meaningful but still not transferable. Prompt shift is especially difficult: leave-one-condition-out correlations fall to about 0.38--0.44. Verbalizer shift transfers better, suggesting that it is closer to a calibration or scale change than prompt shift.

Appendix~\ref{app:alignment} tests a simple distribution-level affine alignment. It does not recover transfer and sometimes makes correlations negative. This negative result supports a limited conclusion: the transfer gap is not merely a trivial mean/scale mismatch in the raw coordinate system. We do not rule out that stronger supervised alignments such as CCA, Procrustes, or nonlinear mappings could recover additional transfer; rather, the result argues against a single zero-shot invariant linear coordinate.

\subsection{Layer and sample-size robustness}

Appendix~\ref{app:splits} repeats readout experiments over ten grouped splits and shows smooth sample-size scaling from 8 to 72 training samples. Appendix~\ref{app:layer} sweeps all 36 layers. The best recovery appears in a broad mid-to-late band centered around layers 20--22; layer 21 is representative, not cherry-picked.

\section{Commitment Is Not Merely Realization Progress}

\subsection{The cursor confound}

A delayed-verdict response confounds two kinds of variation: what answer the model is heading toward and where the model is in the template. The latter is the realization cursor. This is the most natural alternative explanation of the readout results: perhaps $\delta_t$ is recoverable only because hidden states encode template progress. We therefore ask a stronger question: can answer-preference information and cursor information be assigned different predictive roles within a low-dimensional state summary?

\subsection{Operational factorization}

\begin{figure}[t]
\centering
\includegraphics[width=\linewidth]{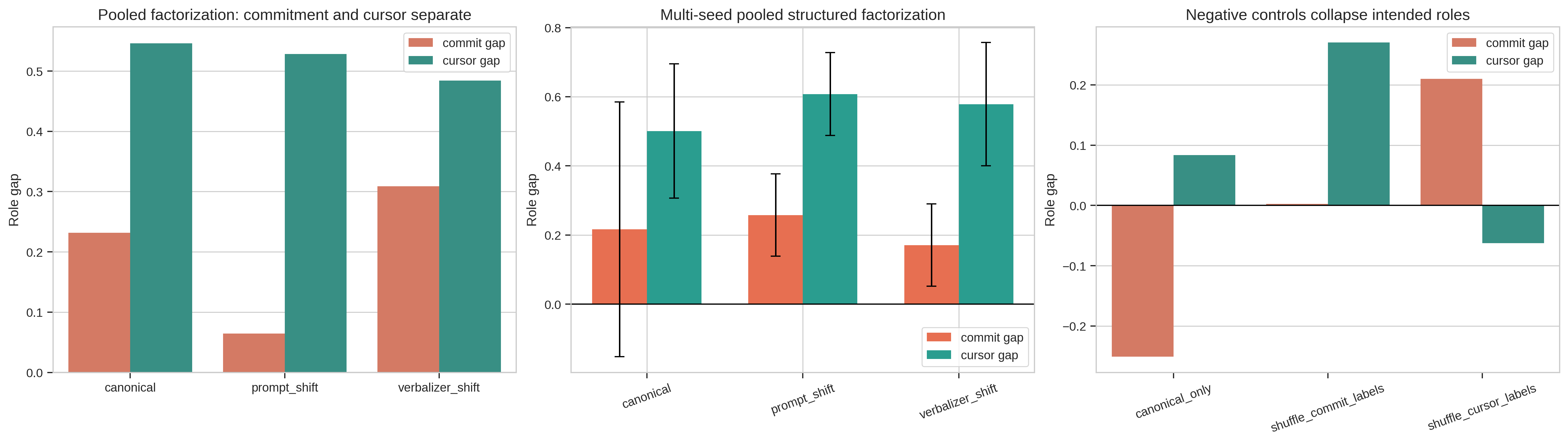}
\caption{\textbf{Commitment can be operationally separated from realization progress.}
The learned factor $u$ is commitment-dominant and $v$ is cursor-dominant under post-hoc probes. Pooled structured factorization yields positive role gaps, the pattern persists over multiple seeds, and negative controls collapse the intended signal. The claim is operational rather than ontological: the factors are validated by their predictive roles, not by uniqueness.}
\label{fig:factorization}
\end{figure}

We train an encoder $E_\phi(q_t)=(u_t,v_t)$, where $q_t$ is a compact hidden summary and $u_t,v_t$ are low-dimensional factors. The factor $u_t$ is trained to be commitment-dominant and $v_t$ cursor-dominant. We evaluate this operationally rather than ontologically: after training, separate post-hoc probes test how well $u$ and $v$ predict $\delta$ and cursor targets. The commitment role gap is the advantage of $u$ over $v$ for predicting $\delta$; the cursor role gap is the advantage of $v$ over $u$ for predicting cursor variables. Leakage is measured by the weaker factor's ability to predict the other target. The goal is not unique disentanglement, but a reproducible separation of predictive roles under multi-seed runs and shuffled-label controls.
\Cref{fig:factorization} shows three pieces of evidence. First, pooled structured factorization produces positive commitment and cursor gaps. Second, these gaps persist across multiple seeds. Third, shuffling the corresponding labels collapses the intended roles. Appendix~\ref{app:factor} reports full multi-seed tables, negative controls, bootstrap intervals, and frozen-encoder reprobe results.

\subsection{Interpretation}

The factorization results support a calibrated claim. We do not claim unique or complete disentanglement: commitment and cursor information can coexist, and canonical-only structured factorization is harder than pooled factorization. The important point is operational. Under the criteria we test, the state admits low-dimensional factors with different predictive roles. This rules out the simplest cursor-only explanation of the commitment readout while avoiding an ontological claim about uniquely separated internal variables.

\section{Robustness Beyond the Main Template}

The main experiments use controlled templates because exact scoring and onset annotation require parser-compatible outputs. Appendix experiments test whether the findings survive beyond this setup. Additional prompt paraphrases, alternative verbalizers, task variants, and a free-form delayed-verdict setting all show positive lead on parsed trajectories and strong low-dimensional recovery. Cross-model replications on Qwen2.5-3B, Qwen2.5-7B, and GLM-4-9B support that the phenomenon is not unique to Qwen3 \citep{qwen2025qwen3,glm2024chatglm}, while also showing that parser compatibility is part of onset-based experimental design. Finally, a stricter quality-gated replication achieves parse rate 1.0 in all four conditions and preserves positive lead, though with shorter magnitude. Together, these diagnostics remove several alternative explanations: cutoff arbitrariness, split artifacts, layer cherry-picking, template specificity, Qwen3 specificity, parsed-subset selection, and trivial affine mismatch in transfer.

\section{Discussion}

The framework replaces an informal psychological phrase with a computable model-internal quantity. Instead of asking whether the model ``knows'' or ``has decided'' in an unrestricted sense, we ask whether the continuation-score projection of its autoregressive state onto a specified finite answer set has stabilized. This is closer to a state-dependent finite-answer forecast, or revealed continuation preference, than to an unrestricted belief claim, aligning with distinctions made in work on probes, latent knowledge, and model self-knowledge \citep{belinkov2022probing,burns2023discovering,kadavath2022language}. In binary tasks, the object is exactly $\delta(\xi)$.

Lead measures how many tokens before explicit answer onset the contextual finite-answer winner retrospectively becomes stable with sufficient margin. In the main templates, this lead is often 17--31 tokens; in the stricter parser-clean replication it remains positive but shorter. This does not prove that post-stabilization text is unfaithful, causally irrelevant, or merely rationalizing. It only shows that the finite-answer projection has already stabilized on the eventual answer while visible generation continues.

The stabilization time is retrospective: naive online rules can stop too early, although calibrated online detectors can work in controlled settings. The projection is verbalizer-conditioned, consistent with prompt-classification and calibration results showing verbalizer sensitivity \citep{schick2021exploiting,gao2021making,zhao2021calibrate}. The representation results also show shared information without a single invariant coordinate; a direction found in one prompt format should not be assumed to transfer unchanged to another.

The main results are observational. A small exact-scoring intervention pilot, related in spirit to activation-editing and causal-intervention work \citep{meng2022locating,geiger2021causal}, shows that residual-stream steering can locally shift $\delta$. However, the shifts are small, winner flips are rare, and generated final answers do not reliably flip. Thus the pilot supports local causal sensitivity of the finite-answer projection, not causal control over final answer generation.

\section{Limitations and Conclusion}

The scalar code $\delta$ applies directly to binary finite-answer tasks; for $K>2$, the projection becomes a score vector modulo an additive constant, and commitment criteria must specify whether stabilization means top-answer persistence, margin to the runner-up, or low entropy over the answer simplex. We leave empirical multi-class evaluation to future work. The measured object is verbalizer-conditioned and should not be interpreted as unrestricted semantic belief. The stabilization time is retrospective rather than an online detector, although calibrated online detectors can be trained in controlled settings. The experiments use controlled delayed-verdict templates and deterministic greedy decoding, which make onset annotation reliable and trajectories reproducible but are simpler than open-ended or stochastic generation. Lead magnitude depends on template geometry, and the causal pilot demonstrates local sensitivity of $\delta$ rather than reliable final-answer control. Under stochastic decoding, commitment could be defined relative to each sampled trajectory or to the distribution over sampled final answers; evaluating this extension is future work.

We introduced a finite-answer framework for measuring pre-verbalization preference stabilization in autoregressive language models. The key object is an exact continuation-score projection of the model's own state onto a specified finite answer set; in binary tasks it reduces to a one-dimensional log-odds code $\delta(\xi)$. In controlled delayed-verdict tasks with Qwen3-4B-Instruct, this contextual finite-answer preference often stabilizes before the final answer is verbalized, tracks the model's eventual output rather than ground truth, is linearly recoverable from hidden states, and is shared across neighboring conditions without forming a single zero-shot invariant direction. Parser-clean replication, online diagnostics, bare/contextual comparisons, and exact-scoring intervention pilots clarify the scope of the claim: this is a precise finite-answer measurement, not a verbalizer-free belief, not automatically an online detector, and not reliable causal control over generation.

\bibliography{iclr2026_conference}
\bibliographystyle{iclr2026_conference}

\newpage

\appendix
\section*{Appendix Overview}

The appendix is organized as a set of checks around the main finite-answer measurement. Mathematical and implementation details come first, followed by full numeric tables and robustness analyses. Later sections report parser diagnostics, quality-gated replication, online and verbalizer diagnostics, and limited causal-sensitivity pilots.

\begin{center}
\begin{tabularx}{\linewidth}{>{\bfseries}lX}
\toprule
Section & Contents \\
\midrule
Appendix~\ref{app:math} & Mathematical details for continuation scores, answer marginalization, binary $\delta$, and greedy-rollout tautology. \\
Appendix~\ref{app:expdetails} & Prompts, parsers, verbalizers, and pipeline sanity checks. \\
Appendix~\ref{app:fulltables} & Full readout and transfer tables for the main Qwen3 experiment. \\
Appendices~\ref{app:threshold}--\ref{app:factor} & Threshold, split, sample-size, layer, and factorization robustness. \\
Appendices~\ref{app:extra}--\ref{app:alignment} & Additional prompt/task/verbalizer settings, cross-model replications, and alignment-lite analysis. \\
Appendices~\ref{app:parser_diag}--\ref{app:quality_gated} & Parser diagnostics and quality-gated parser-clean replication. \\
Appendices~\ref{app:causal_pilot}--\ref{app:head_exploratory} & Exact-scoring causal-sensitivity pilot and exploratory head-level diagnostic. \\
Appendix~\ref{app:trajectory_anatomy} & Representative and aggregate trajectory-level anatomy. \\
Appendices~\ref{app:additional_diagnostics}--\ref{app:bare_contextual} & Correctness, cursor, online-detection, and bare/contextual verbalizer diagnostics. \\
Appendix~\ref{app:repro} & Reproducibility checklist. \\
\bottomrule
\end{tabularx}
\end{center}

\section{Mathematical Details}
\label{app:math}

This appendix expands the mathematical definitions used in the main paper. Its role is to make explicit that finite-answer commitment is an exact continuation-score object, not a learned probe target.

\subsection{Sequence score equals conditional log-probability}

Let $\xii_n$ be induced by prefix $x_{1:n}$, and let $y_{1:T}$ be a continuation. By autoregressive factorization,
\[
p_\theta(y_{1:T}\mid x_{1:n})
=
\prod_{t=1}^T
p_\theta(y_t\mid x_{1:n},y_{1:t-1}).
\]
Let $\xii^y_{t-1}$ be the state after appending $y_{1:t-1}$ to $\xii_n$. Then
\[
p_\theta(y_t\mid x_{1:n},y_{1:t-1})
=
\Pi_\theta(\xii^y_{t-1})_{y_t}.
\]
Taking logs gives
\[
\log p_\theta(y_{1:T}\mid x_{1:n})
=
\sum_{t=1}^T
\log \Pi_\theta(\xii^y_{t-1})_{y_t}
=
S_\theta(y\mid\xii_n).
\]

\subsection{Answer-level marginalization}

For answer $a$ with verbalizer set $\Sigma(a)$,
\[
P_\theta(\Sigma(a)\mid\xii)
=
\sum_{y\in\Sigma(a)}
p_\theta(y\mid\xii)
=
\sum_{y\in\Sigma(a)}
\exp S_\theta(y\mid\xii).
\]
Therefore
\[
S_\theta(a\mid\xii)
=
\log P_\theta(\Sigma(a)\mid\xii)
=
\log\sum_{y\in\Sigma(a)}\exp S_\theta(y\mid\xii).
\]

\subsection{Binary sufficiency of $\delta$}

Let $s_y=S_\theta(\yes\mid\xii)$ and $s_n=S_\theta(\no\mid\xii)$. Then
\[
\Ptilde(\yes\mid\xii)
=
\frac{e^{s_y}}{e^{s_y}+e^{s_n}}
=
\frac{1}{1+e^{-(s_y-s_n)}}.
\]
Thus $\Ptilde(\yes\mid\xii)=\sigma(\delta(\xi))$, and conversely
\[
\delta(\xi)
=
\log\frac{\Ptilde(\yes\mid\xii)}
{1-\Ptilde(\yes\mid\xii)}.
\]
So $\delta$ and the binary induced answer distribution are equivalent parameterizations.

\subsection{Commitment time in sign notation}

Let
\[
w_t=
\begin{cases}
+1,&\delta_t\ge0,\\
-1,&\delta_t<0,
\end{cases}
\qquad
w^\star=
\begin{cases}
+1,&a^\star=\yes,\\
-1,&a^\star=\no.
\end{cases}
\]
Then
\[
\taucommit
=
\min
\left\{
t<\tauonset:
w_t=w^\star,\;
|\delta_t|\ge\gamma,\;
w_s=w^\star\;\forall s\in[t,\tauonset)
\right\}.
\]
This definition rules out transient agreement followed by a later flip.

\subsection{Greedy rollout constancy}

Along a deterministic greedy trajectory, token $r_{t+1}$ satisfies
\[
r_{t+1}=\arg\max_u \Pi_\theta(\xi_t)_u.
\]
Starting greedy decoding from $\xi_t$ again selects $r_{t+1}$, reaches $\xi_{t+1}$, and inductively reproduces the same suffix. Thus a constant greedy final answer along a greedy trajectory is not evidence of commitment. It is a tautology of deterministic decoding.

\section{Experimental Details}
\label{app:expdetails}

This appendix documents prompts, parsers, verbalizers, hidden-state extraction, cursor targets, and scoring details. Its purpose is to make the exact target reproducible.

\subsection{Canonical parity prompt}

\begin{verbatim}
Fill the exact 5-line template and nothing else.

s1 = a + b = ?
s2 = c + d = ?
total = s1 + s2 = ?
parity = even or odd
Verdict: yes or no

Question: Is a+b+c+d even?
a=..., b=..., c=..., d=...

Rules:
- Keep exactly 5 lines.
- Use the same left-hand labels.
- The last line must be exactly "Verdict: yes" or "Verdict: no".
\end{verbatim}

\subsection{Prompt shift}

\begin{verbatim}
Write exactly five lines, no extra text.

line1: left_part = a+b = ?
line2: right_part = c+d = ?
line3: sum_all = left_part+right_part = ?
line4: even_check = even or odd
line5: Final answer = yes or no

Task: decide whether a+b+c+d is even.
Values: a=..., b=..., c=..., d=...

The fifth line must be exactly "Final answer = yes" or
"Final answer = no".
\end{verbatim}

\subsection{Verbalizer shift}

The final line is:
\begin{verbatim}
Decision: affirmative or negative
\end{verbatim}
We map affirmative to yes and negative to no.

\subsection{Task-family shift}

The comparison task asks whether $(a+b)>(c+d)$ and ends with \texttt{Verdict: yes/no}. The reasoning template computes left sum, right sum, gap, and comparison.

\subsection{Parsers}

Canonical parser:
\[
\texttt{\string^ \textbackslash s*Verdict:\textbackslash s*(yes|no)\textbackslash s*\$}.
\]
Prompt-shift parser:
\[
\texttt{\string^ \textbackslash s*Final answer\textbackslash s*[:=]\textbackslash s*(yes|no)\textbackslash s*\$}.
\]
Verbalizer-shift parser maps \texttt{Decision: affirmative} to yes and \texttt{Decision: negative} to no. Cross-model relaxed prompt-shift parsing additionally accepts line-numbered variants such as \texttt{line5: Final answer = yes}.

\subsection{Sanity checks}
\label{app:sanity}

\begin{table}[h]
\centering
\caption{\textbf{Pipeline sanity checks.}
These checks rule out common implementation artifacts in generation, scoring, and parsing.}
\label{tab:app_sanity}
\tabfit{
\begin{tabular}{lc}
\toprule
Check & Result \\
\midrule
Custom greedy generation matches HF \texttt{generate} & 1.000 \\
Parser freeze rate & 1.000 \\
Teacher-forced score consistency absolute error & 0.125 \\
\bottomrule
\end{tabular}
}
\end{table}

\section{Full Main Tables}
\label{app:fulltables}

This appendix provides fuller numeric summaries for the main Qwen3 experiment. The main paper reports the numbers needed for the central claims; here we retain additional metrics.

\subsection{Within-condition and pooled readouts}

\begin{table}[h]
\centering
\caption{\textbf{Low-dimensional recovery summary over grouped splits.}
Correlations are averaged over ten splits. The high values show that the exact commitment code is recoverable from compact hidden-state summaries.}
\label{tab:app_lowdim_summary}
\tabfit{
\begin{tabular}{llllrrrr}
\toprule
Feature & Eval type & Train & Test & Corr. mean & Corr. std & Acc.$|\delta|\ge5$ & Tau MAE \\
\midrule
last\_L21 & within & canonical & canonical & 0.955 & 0.009 & 0.991 & 5.17 \\
last\_L21 & within & prompt & prompt & 0.950 & 0.005 & 0.988 & 6.53 \\
last\_L21 & within & verbalizer & verbalizer & 0.929 & 0.019 & 0.982 & 4.17 \\
concat & within & canonical & canonical & 0.959 & 0.009 & 0.995 & 5.17 \\
concat & within & prompt & prompt & 0.954 & 0.006 & 0.989 & 7.27 \\
concat & within & verbalizer & verbalizer & 0.936 & 0.017 & 0.985 & 3.25 \\
\midrule
last\_L21 & pooled & pooled3 & canonical & 0.955 & 0.006 & 0.995 & 4.82 \\
last\_L21 & pooled & pooled3 & prompt & 0.947 & 0.005 & 0.990 & 7.07 \\
last\_L21 & pooled & pooled3 & verbalizer & 0.936 & 0.010 & 0.986 & 5.15 \\
concat & pooled & pooled3 & canonical & 0.923 & 0.009 & 0.986 & 6.40 \\
concat & pooled & pooled3 & prompt & 0.913 & 0.005 & 0.978 & 8.27 \\
concat & pooled & pooled3 & verbalizer & 0.910 & 0.013 & 0.977 & 5.76 \\
\bottomrule
\end{tabular}
}
\end{table}

\subsection{Transfer}

\begin{table}[h]
\centering
\caption{\textbf{Transfer summary.}
Leave-one-condition-out and canonical transfer are weaker than pooled training, especially under prompt shift. This supports the condition-dependent-coordinate interpretation.}
\label{tab:app_transfer}
\tabfit{
\begin{tabular}{llllrrr}
\toprule
Feature & Eval type & Train & Test & Corr. mean & Acc.$|\delta|\ge5$ & Tau MAE \\
\midrule
last\_L21 & LOCO & others & canonical & 0.770 & 0.965 & 8.80 \\
last\_L21 & LOCO & others & prompt & 0.444 & 0.701 & 14.77 \\
last\_L21 & LOCO & others & verbalizer & 0.728 & 0.877 & 9.50 \\
concat & LOCO & others & canonical & 0.704 & 0.949 & 10.31 \\
concat & LOCO & others & prompt & 0.378 & 0.673 & 19.98 \\
concat & LOCO & others & verbalizer & 0.698 & 0.783 & 10.91 \\
\midrule
last\_L21 & canonical transfer raw & canonical & prompt & 0.448 & 0.716 & 14.13 \\
last\_L21 & canonical transfer affine & canonical & prompt & 0.448 & 0.782 & 15.85 \\
last\_L21 & canonical transfer raw & canonical & verbalizer & 0.742 & 0.815 & 9.63 \\
last\_L21 & canonical transfer affine & canonical & verbalizer & 0.742 & 0.939 & 11.29 \\
\bottomrule
\end{tabular}
}
\end{table}

\section{Grouped Bootstrap Confidence Intervals}
\label{app:bootstrap}

Token states within a trajectory are not independent. All confidence intervals in this appendix are therefore computed by grouped bootstrap over trajectories. \Cref{tab:app_bootstrap} reports the main grouped-bootstrap estimates.

\begin{table}[h]
\centering
\caption{\textbf{Grouped bootstrap confidence intervals for main metrics.}
Intervals are 95\% bootstrap intervals over trajectories.}
\label{tab:app_bootstrap}
\tabfit{
\begin{tabular}{lrrrr}
\toprule
Condition & Accuracy & Commit rate & Mean lead & Mean onset \\
\midrule
Canonical & 1.000 [1.000, 1.000] & 1.000 [1.000, 1.000] & 17.41 [14.90, 20.19] & 43.09 [42.24, 44.18] \\
Prompt shift & 1.000 [1.000, 1.000] & 1.000 [1.000, 1.000] & 19.69 [16.61, 22.83] & 53.76 [51.84, 55.53] \\
Task-family shift & 0.844 [0.750, 0.922] & 0.969 [0.922, 1.000] & 31.10 [28.33, 33.60] & 36.55 [36.00, 37.55] \\
Verbalizer shift & 1.000 [1.000, 1.000] & 1.000 [1.000, 1.000] & 17.19 [14.58, 20.05] & 41.25 [40.76, 42.03] \\
\bottomrule
\end{tabular}
}
\end{table}

\section{Retrospective Stabilization Time}
\label{app:threshold}

This appendix tests whether the main lead result depends on the threshold $\gamma=2$. It does not: increasing $\gamma$ delays commitment but does not remove positive lead.

\begin{figure}[h]
\centering
\includegraphics[width=\linewidth]{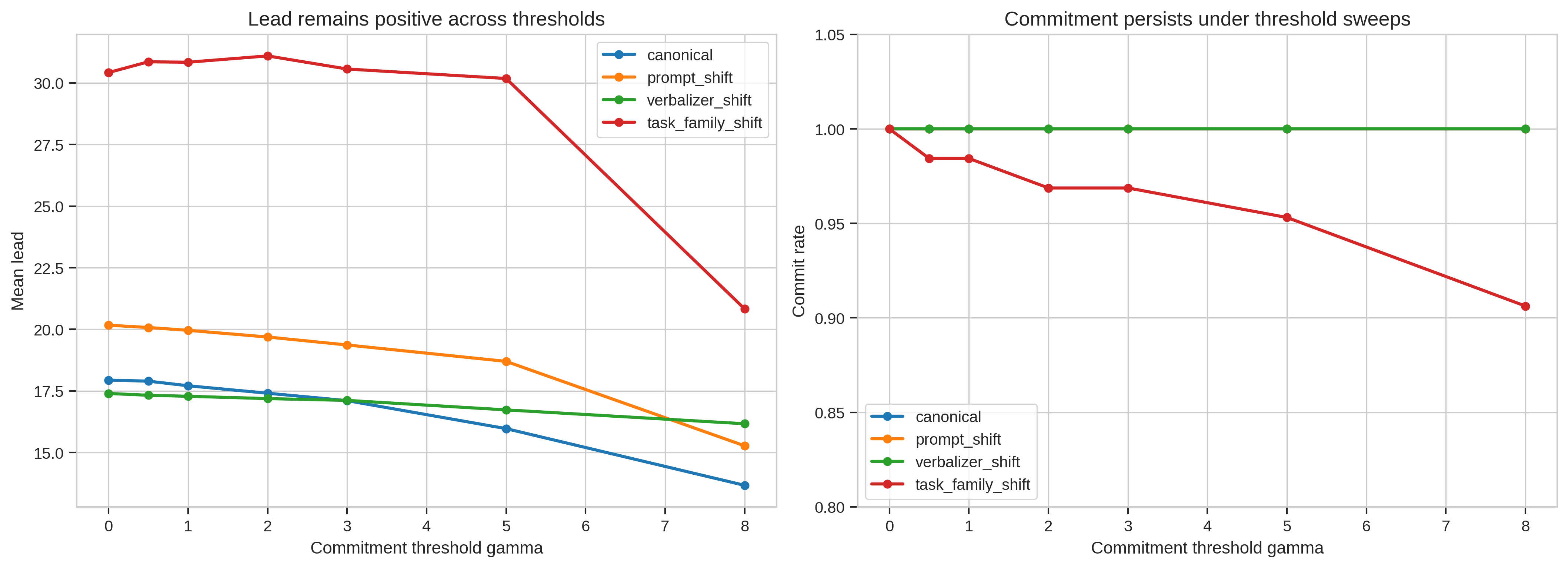}
\caption{\textbf{Retrospective stabilization time of commitment.}
Left: mean lead remains positive across all tested commitment thresholds. Right: commitment persists under threshold sweeps, with only mild decay in commit rate for the hardest task-family-shift condition at large $\gamma$.}
\label{fig:app_gamma}
\end{figure}

At $\gamma=8$, mean lead remains positive in all main conditions: canonical 13.67, prompt shift 15.27, verbalizer shift 16.17, and task-family shift 20.83. Thus the phenomenon is not a cutoff artifact.

\section{Split and Sample-Size Robustness}
\label{app:splits}

This appendix tests whether low-dimensional recoverability is an artifact of a single split or training set size.

\begin{figure}[h]
\centering
\includegraphics[width=\linewidth]{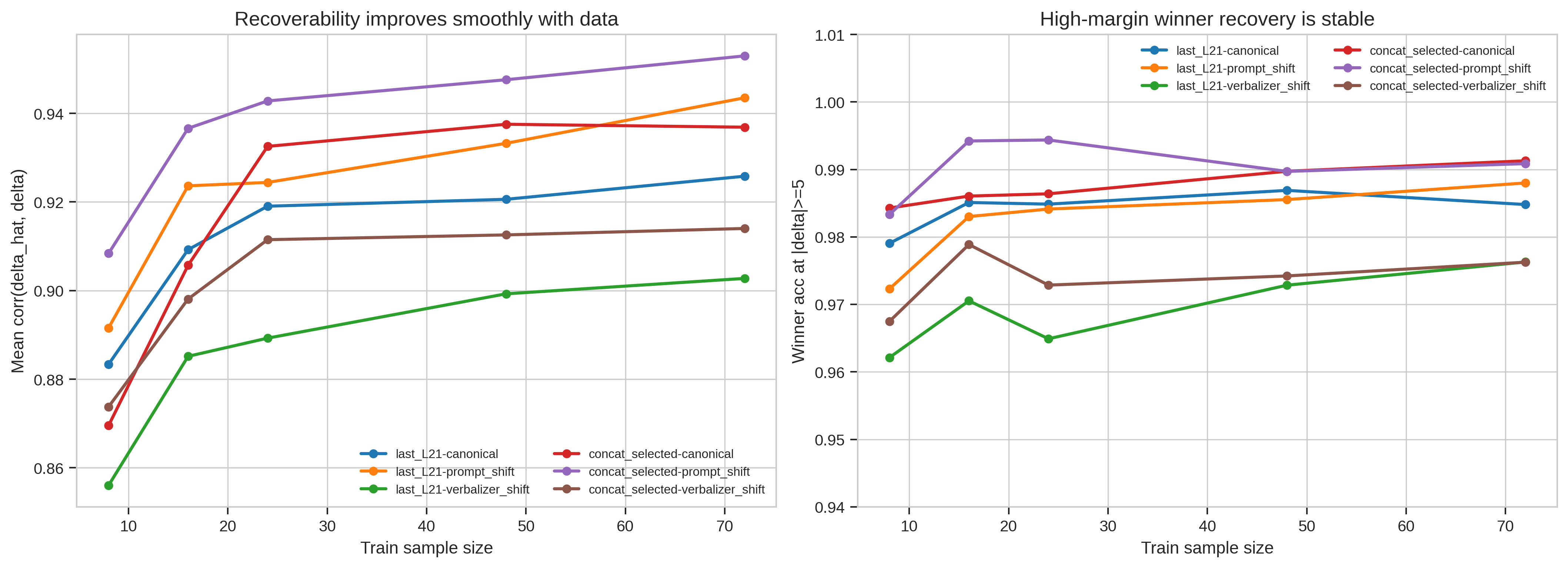}
\caption{\textbf{Sample-size scaling of commitment recovery.}
Left: $\delta$-correlation improves smoothly as training sample size increases. Right: high-margin winner recovery remains stable across training sizes.}
\label{fig:app_samplesize}
\end{figure}

Recovery is already strong at 8 samples and improves smoothly up to 72 samples. Across ten grouped splits, standard deviations are small, especially for pooled and within-condition settings.

\section{Layer Sweep}
\label{app:layer}

This appendix tests whether layer 21 is a cherry-picked readout layer. It is not: commitment recovery is strongest in a broad mid-to-late band centered around layers 20--22.

\begin{figure}[h]
\centering
\includegraphics[width=\linewidth]{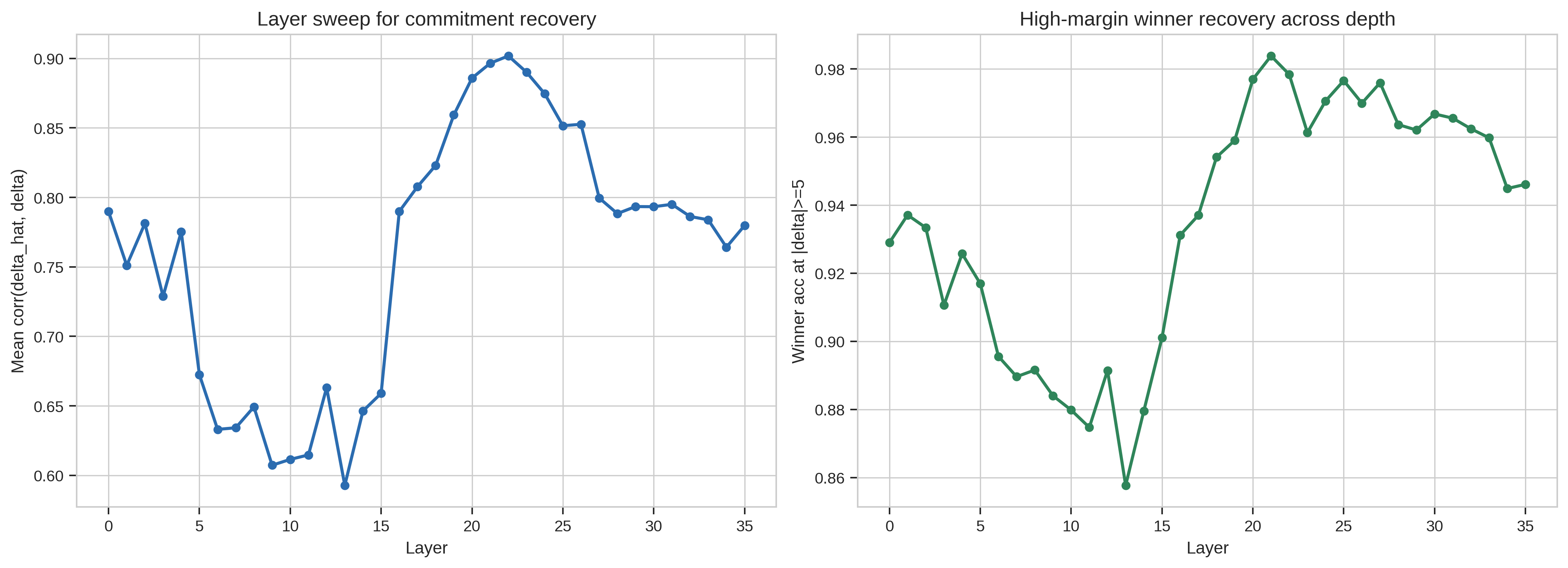}
\caption{\textbf{Layer sweep for commitment recovery.}
The best recoverability appears in a broad mid-to-late band rather than at a single isolated layer. Layer 21 is representative but not unique.}
\label{fig:app_layer}
\end{figure}

The top layers by mean $\delta$-correlation are layer 22 (0.902), layer 21 (0.897), layer 23 (0.890), and layer 20 (0.886). This supports using \texttt{last\_L21} as a compact anchor.

\section{Factorization Robustness}

\subsection{Factorization objective and metrics}

Let $q_t$ denote the hidden summary at pre-onset state $t$. The encoder maps $q_t$ to two factors,
\[
E_\phi(q_t)=(u_t,v_t), \qquad u_t,v_t\in\mathbb{R}^8.
\]
The intended roles are asymmetric: $u_t$ should preserve finite-answer preference information, while $v_t$ should preserve realization-progress information. We train the structured factorization with supervised losses for the intended targets and evaluate the result with post-hoc probes.

For reporting, we define the commitment gap as
\[
\mathrm{Gap}_{\delta}
=
\mathrm{Perf}(u\rightarrow \delta)
-
\mathrm{Perf}(v\rightarrow \delta),
\]
where performance is the held-out correlation for continuous $\delta$ prediction. We define the cursor gap as
\[
\mathrm{Gap}_{c}
=
\mathrm{Perf}(v\rightarrow c)
-
\mathrm{Perf}(u\rightarrow c),
\]
where $c$ denotes the cursor target and performance is the corresponding held-out predictive score. Leakage is the non-dominant factor's predictive performance on the other target. A successful operational separation requires positive role gaps, limited leakage, and collapse under shuffled-target controls.

\label{app:factor}

This appendix expands the factorization analysis. Its purpose is to test whether the operational separation between commitment and cursor depends on one seed or one split.

\begin{table}[h]
\centering
\caption{\textbf{Multi-seed factorization summary.}
Positive commitment and cursor gaps persist across pooled structured settings.}
\label{tab:app_factor_ms}
\tabfit{
\begin{tabular}{llllrrrr}
\toprule
Feature & Cursor & Train & Test & $u\to\delta$ & $v\to\delta$ & Commit gap & Cursor gap \\
\midrule
last\_L21 & structured & pooled3 & canonical & 0.939 & 0.723 & 0.217 & 0.501 \\
last\_L21 & structured & pooled3 & prompt & 0.817 & 0.559 & 0.258 & 0.608 \\
last\_L21 & structured & pooled3 & verbalizer & 0.788 & 0.617 & 0.171 & 0.578 \\
concat & structured & pooled3 & canonical & 0.818 & 0.401 & 0.418 & 0.699 \\
concat & structured & pooled3 & prompt & 0.914 & 0.606 & 0.308 & 0.763 \\
concat & structured & pooled3 & verbalizer & 0.822 & 0.532 & 0.290 & 0.850 \\
\bottomrule
\end{tabular}
}
\end{table}

The most stable claim is the pooled structured one: $u$ tends to be commitment-dominant and $v$ cursor-dominant. Canonical-only structured factorization is more variable, which we interpret as a harder regime rather than a refutation.

\section{Additional Prompt, Verbalizer, Task, and Free-Form Robustness}
\label{app:extra}

This appendix tests whether the phenomenon is tied to the main templates. It is not: additional prompt paraphrases, task variants, verbalizers, and a free-form delayed-verdict setting reproduce the same qualitative pattern.

\begin{figure}[h]
\centering
\includegraphics[width=\linewidth]{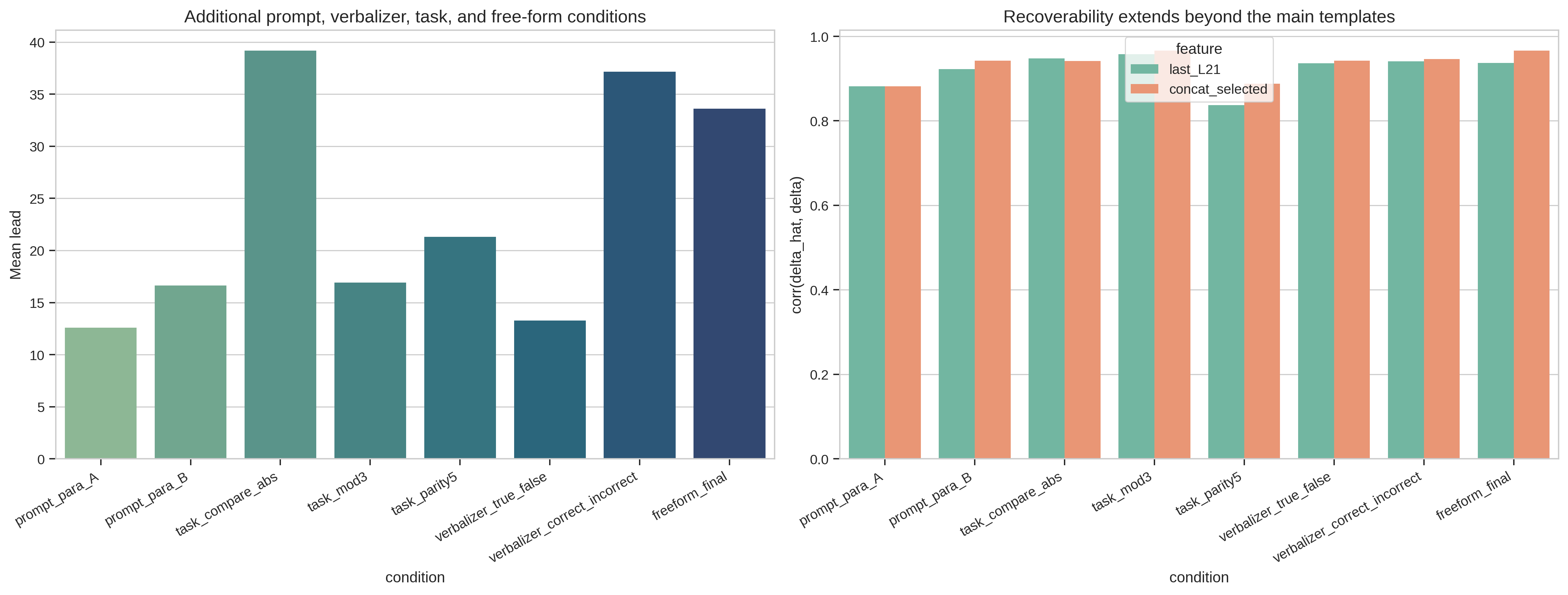}
\caption{\textbf{Additional prompt, verbalizer, task, and free-form conditions.}
Left: mean lead remains positive across all additional settings. Right: exact commitment recovery remains strong across the same settings.}
\label{fig:app_extra}
\end{figure}

Prompt paraphrases have parse rate 1.0, sample accuracy 1.0, and mean lead 12.59--16.63. Task variants have parse rate 1.0, sample accuracy 1.0, and mean lead 16.91--39.19. The free-form setting has lower parse rate but still shows strong commitment and recovery on the parsed subset.

\section{Cross-Model Replications}
\label{app:crossmodel}

This appendix tests whether the phenomenon is unique to Qwen3. It is not, although parser compatibility matters.

\begin{figure}[h]
\centering
\includegraphics[width=\linewidth]{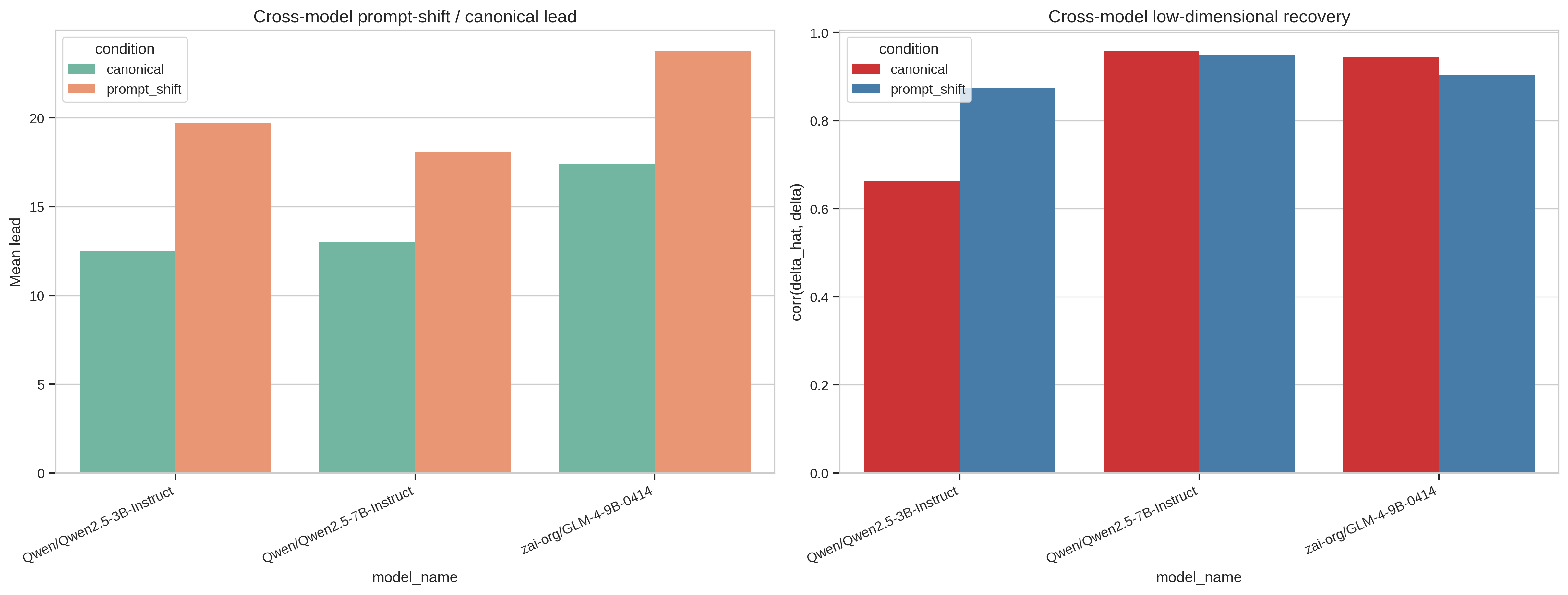}
\caption{\textbf{Cross-model replications.}
Left: mean lead in canonical and prompt-shift conditions for Qwen2.5-3B, Qwen2.5-7B, and GLM-4-9B. Right: low-dimensional recovery on the same models. Strict prompt-shift parsing fails for some models, but relaxed parsing recovers the phenomenon.}
\label{fig:app_cross}
\end{figure}

Qwen2.5-7B reproduces canonical and prompt-shift conditions under strict parsing. Qwen2.5-3B and GLM-4-9B reproduce canonical under strict parsing, and prompt shift under relaxed parsing \citep{qwen2025qwen3,glm2024chatglm}. The relaxed parser accepts common line-numbered variants such as \texttt{line5: Final answer = yes}. This shows that strict parse failures were parser compatibility issues rather than absence of commitment.

\section{Alignment-Lite Analysis}
\label{app:alignment}

This appendix tests whether zero-shot transfer failure is merely a scale or distribution mismatch. A simple distribution-level affine alignment does not recover transfer.

\begin{figure}[h]
\centering
\includegraphics[width=\linewidth]{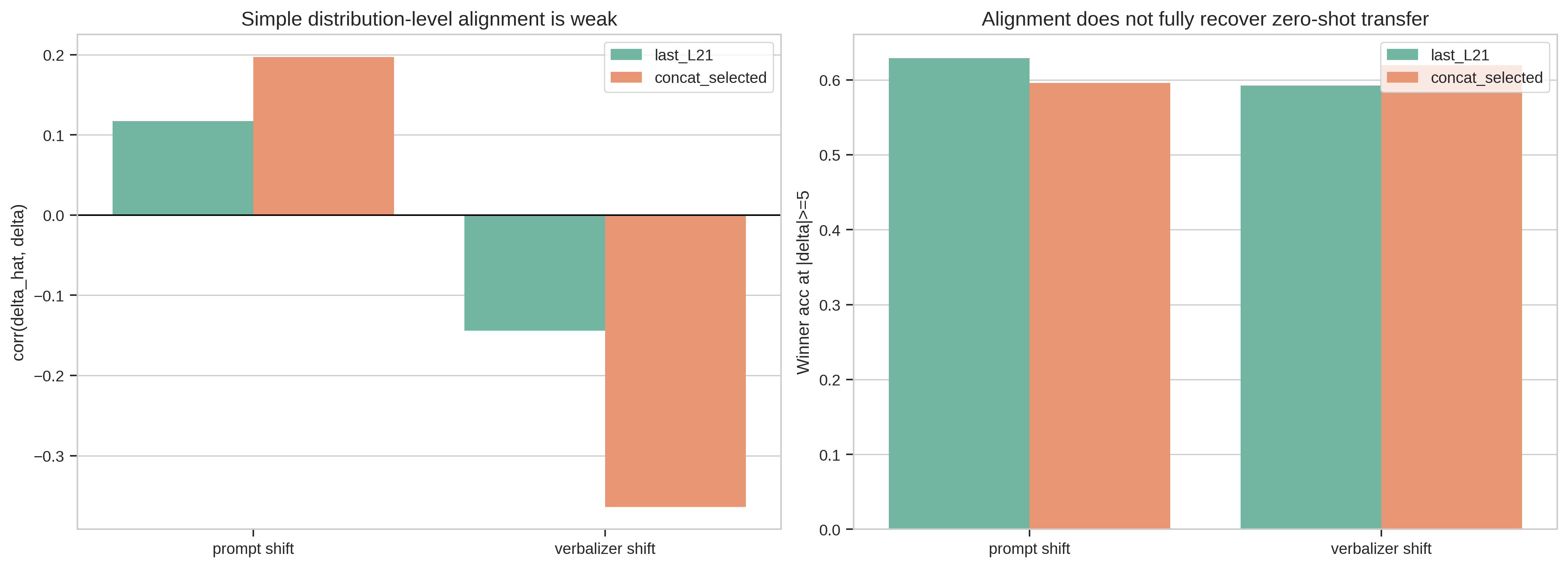}
\caption{\textbf{Alignment-lite analysis.}
Simple distribution-level affine alignment does not recover zero-shot transfer: prompt-shift and verbalizer-shift correlations remain weak, and some cases become negative.}
\label{fig:app_align}
\end{figure}

This negative result supports the interpretation that prompt and verbalizer shifts alter commitment coordinates more structurally than a simple affine distribution correction can capture.

\section{Exact-Scoring Causal-Sensitivity Pilot}
\label{app:causal_pilot}

The main measurement is observational. As a limited causal-sensitivity pilot, inspired by activation editing and causal-intervention approaches \citep{meng2022locating,geiger2021causal}, we intervene on the last-position residual stream and recompute the same exact contextual finite-answer score $\delta$ used in the main analysis. This is stronger than a one-token logit diagnostic because the endpoint remains the full contextual verbalizer projection.

\begin{table}[h]
\centering
\caption{\textbf{Exact residual-stream causal-sensitivity pilot.}
We steer the last-position residual stream and recompute the same exact contextual finite-answer score $\delta$ used in the main analysis. Layer-24 ridge steering shows the clearest dose-response. The effect shifts $\delta$ locally but rarely flips the finite-score winner and does not yield generated-answer control.}
\label{tab:app_v6_causal_pilot}
\tabfit{
\begin{tabular}{lrrrrr}
\toprule
Intervention & $\alpha$ & Mean intended shift & Median intended shift & Frac. moved & Winner flip \\
\midrule
Layer 24 ridge & 1 & 0.0166 & 0.0177 & 0.609 & 0.004 \\
Layer 24 ridge & 2 & 0.0223 & 0.0293 & 0.629 & 0.008 \\
Layer 24 ridge & 4 & 0.0583 & 0.0690 & 0.676 & 0.008 \\
Layer 24 ridge & 8 & 0.1152 & 0.1602 & 0.680 & 0.016 \\
\bottomrule
\end{tabular}
}
\end{table}

\begin{figure}[h]
\centering
\includegraphics[width=0.82\linewidth]{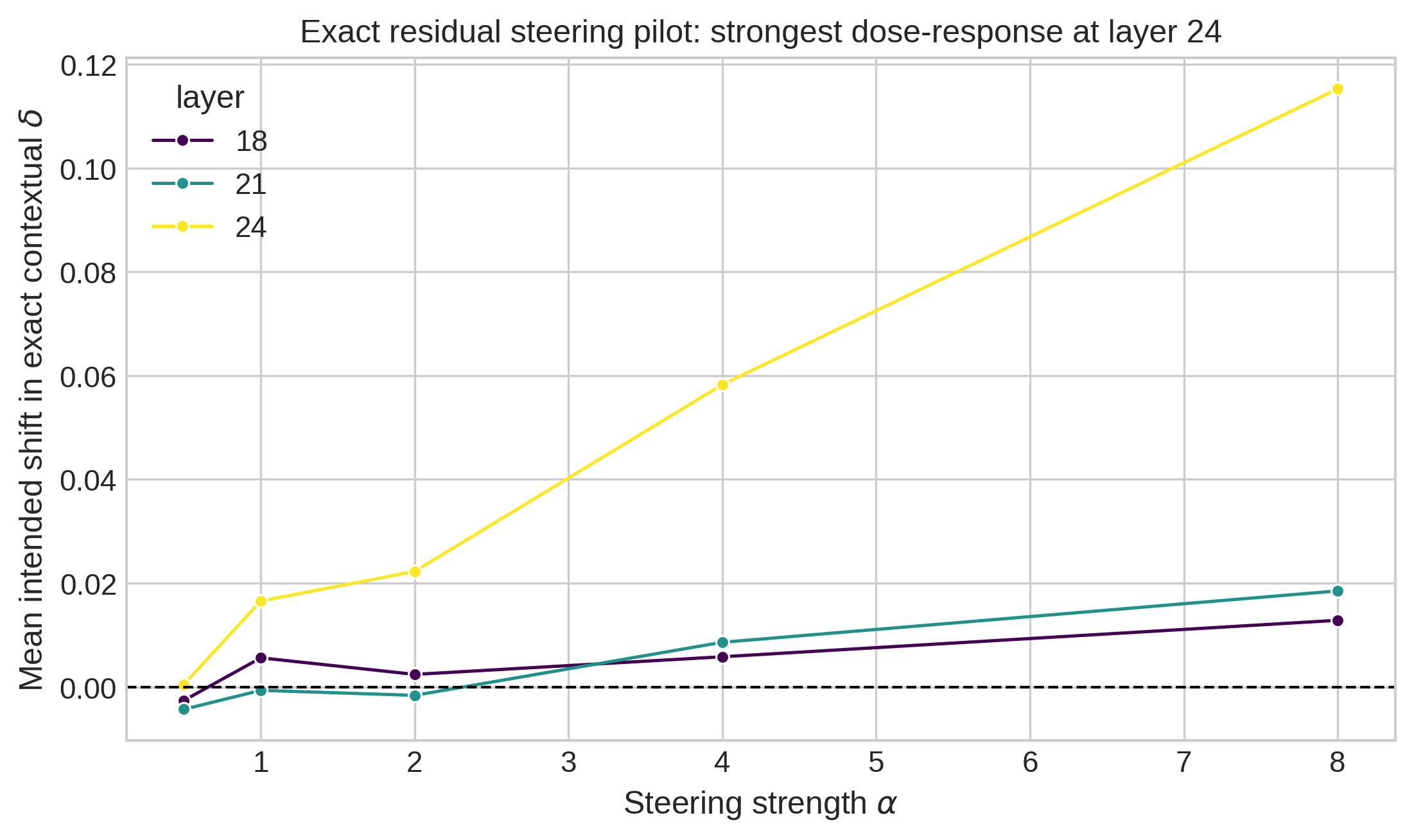}
\caption{\textbf{Exact residual-stream causal-sensitivity pilot.}
Steering along ridge preference directions produces the clearest positive dose-response at layer 24. The effect shifts the exact contextual finite-answer projection locally, but it is small relative to the pre-intervention margin and does not reliably flip generated final answers.}
\label{fig:app_v6_causal_pilot}
\end{figure}

The strongest effect is layer-24 ridge steering. At the largest tested strength, 68\% of states move in the intended direction, with mean intended shift 0.115 in exact contextual $\delta$. However, the mean pre-intervention margin is about 2.58, so the average shift is small relative to the distance from the decision boundary. Finite-score winner flips are rare, and a small generation endpoint preserves parser validity but produces no parsed final-answer flips. We therefore interpret this pilot as evidence of local causal sensitivity, not as causal control over answer generation. A stronger causal account would require systematic patching or ablation across timepoints, especially comparing states before and after $\taucommit$, while controlling for format preservation. We leave this to future work because the present paper's main contribution is the exact measurement target and its temporal behavior.

\section{Exploratory Head-Level Intervention Diagnostic}
\label{app:head_exploratory}

We also explored whether the finite-answer preference signal localizes cleanly to a small set of attention heads. We first screened all attention heads by how well their pre-output-projection head outputs linearly predict exact contextual $\delta$ on training states. Many heads are highly predictive: the maximum screening correlations are 0.969 for canonical, 0.913 for prompt shift, 0.890 for task-family shift, and 0.939 for verbalizer shift.

However, held-out interventions do not cleanly support a localized top-head mechanism. Top readout heads do not consistently outperform random ridge-controlled heads.

\begin{table}[h]
\centering
\caption{\textbf{Exploratory head-level intervention diagnostic.}
Top readout heads do not cleanly outperform random ridge-controlled heads. Thus this experiment does not establish a localized attention-head mechanism; it suggests that finite-answer preference information is distributed and that readout strength is not the same as causal sensitivity.}
\label{tab:app_head_exploratory}
\tabfit{
\begin{tabular}{lrrrrr}
\toprule
Arm & $\alpha$ & Mean intended shift & Median intended shift & Frac. moved & Winner flip \\
\midrule
Random head + ridge & 16 & 0.0701 & 0.0246 & 0.634 & 0.010 \\
Top head + random dir. & 16 & -0.0201 & -0.0016 & 0.488 & 0.010 \\
Top head + ridge & 16 & 0.0263 & 0.0256 & 0.597 & 0.009 \\
\bottomrule
\end{tabular}
}
\end{table}

This result is informative but not a positive mechanism claim. It suggests that finite-answer preference information is distributed across many heads and that linear readout strength is not identical to causal sensitivity. We therefore do not use the head-level experiment as evidence for a localized attention-head circuit.

\section{Trajectory-Level Anatomy}
\label{app:trajectory_anatomy}

The main text reports aggregate lead. This appendix examines the trajectory-level and distributional structure of commitment. Its purpose is to rule out two alternative explanations: first, that the mean lead is driven by a small number of outliers; second, that commitment is a transient spike rather than a stable pre-onset state.

\subsection{Representative trajectory}

\begin{figure}[h]
\centering
\includegraphics[width=\linewidth]{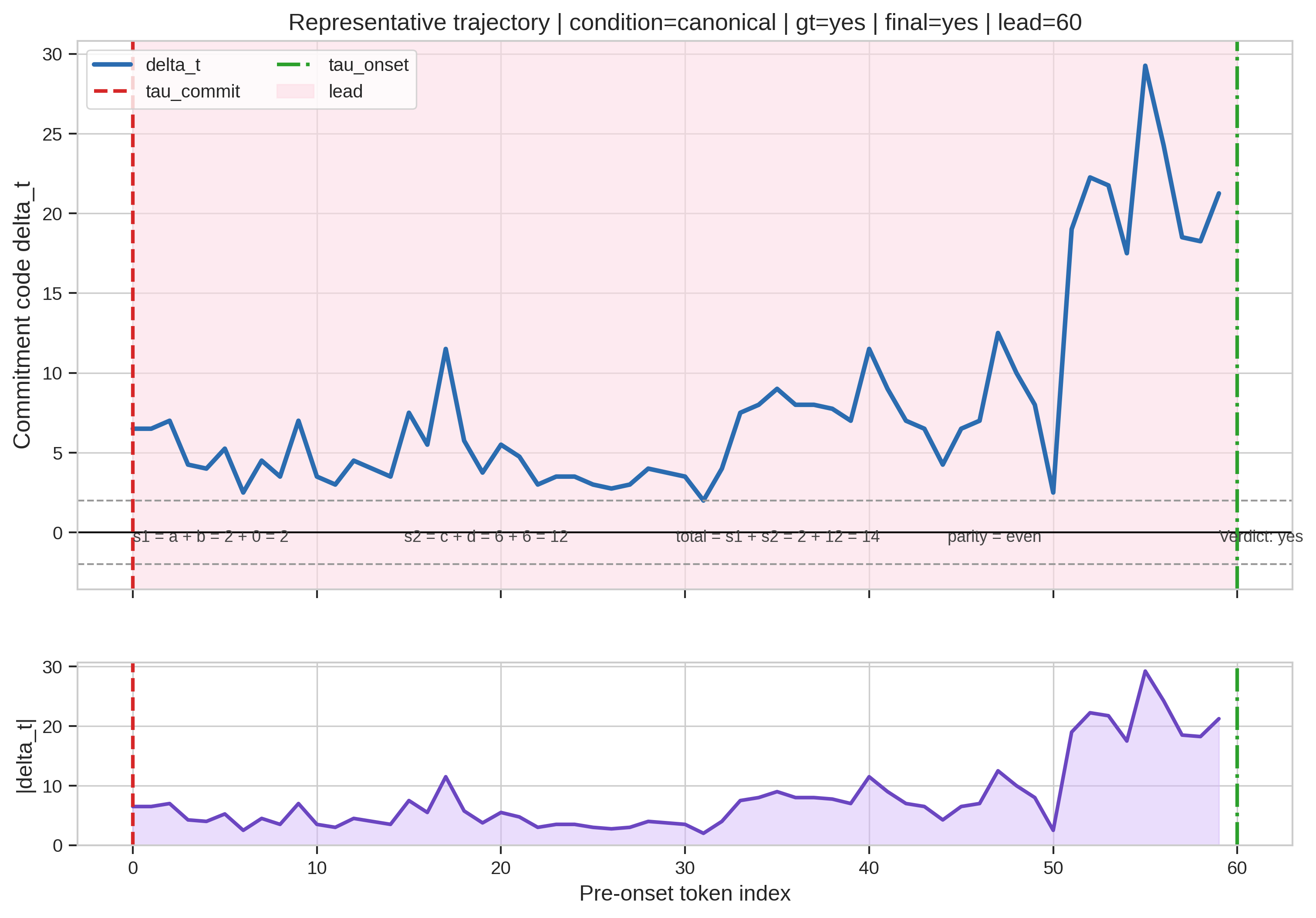}
\caption{\textbf{A trajectory-level view of pre-verbalization commitment.}
For a representative high-lead canonical example, the finite-answer commitment code becomes high-margin and remains aligned with the eventual answer well before the final verdict is parsed. The shaded region is the lead interval between commitment time and answer onset.}
\label{fig:app_representative_trajectory}
\end{figure}

\subsection{Lead distributions}
\label{app:lead_distribution}

\begin{figure}[h]
\centering
\includegraphics[width=0.78\linewidth]{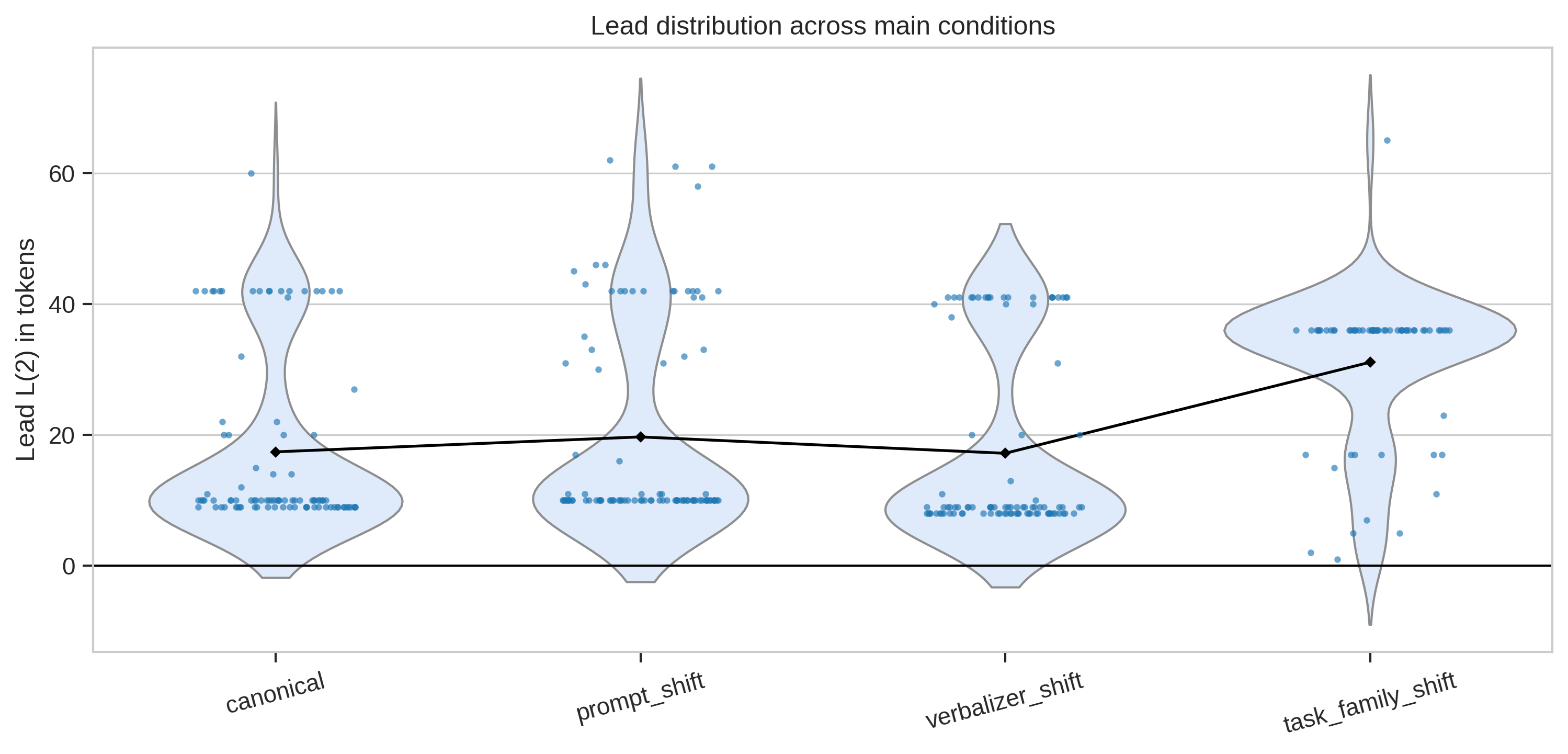}
\caption{\textbf{Lead distributions across main conditions.}
Each point is a parsed trajectory with a defined commitment time at $\gamma=2$. The violin shape shows the full distribution and the black diamond indicates the mean. Positive lead is not driven only by a few extreme cases: all main conditions show broad distributions of pre-verbalization commitment.}
\label{fig:app_lead_distribution}
\end{figure}

The main paper reports mean lead, but the distribution matters. \Cref{fig:app_lead_distribution} shows that positive lead is not merely an average produced by a few extreme trajectories. The mass of the distribution lies above zero in all main conditions. This supports the interpretation that pre-verbalization commitment is a common trajectory-level phenomenon rather than a rare event.

\subsection{Commitment time versus answer onset}

\begin{figure}[h]
\centering
\includegraphics[width=0.62\linewidth]{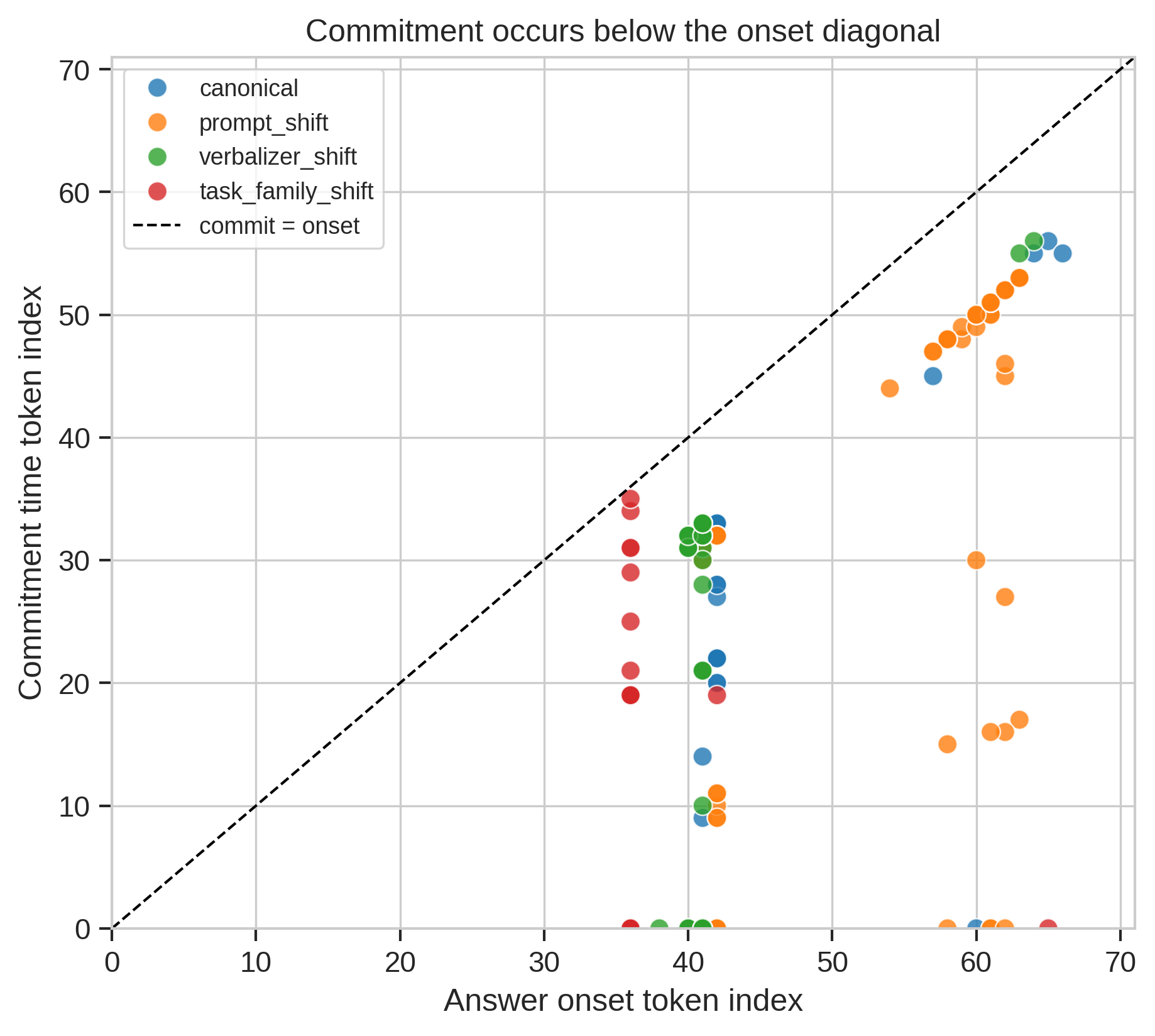}
\caption{\textbf{Commitment time versus answer onset.}
Each point is one committed trajectory. Points below the dashed diagonal have positive lead. The concentration below the diagonal shows that commitment generally occurs before the answer becomes parseable.}
\label{fig:app_onset_commit_scatter}
\end{figure}

\Cref{fig:app_onset_commit_scatter} gives a sample-level view of the same phenomenon. If commitment occurred only when the final answer was written, points would lie near the diagonal. Instead, trajectories concentrate below the diagonal, indicating that commitment typically precedes answer onset.

\subsection{Normalized signed-delta trajectories}

For each trajectory, define the signed commitment code toward the eventual answer:
\[
\delta^{\star}_t =
\begin{cases}
\delta_t, & a^\star=\mathrm{yes},\\
-\delta_t, & a^\star=\mathrm{no}.
\end{cases}
\]
Thus positive values always indicate evidence toward the final parsed answer. We normalize pre-onset time by $t/\tau_{\rm onset}$ and aggregate trajectories within condition.

\begin{figure}[h]
\centering
\includegraphics[width=0.78\linewidth]{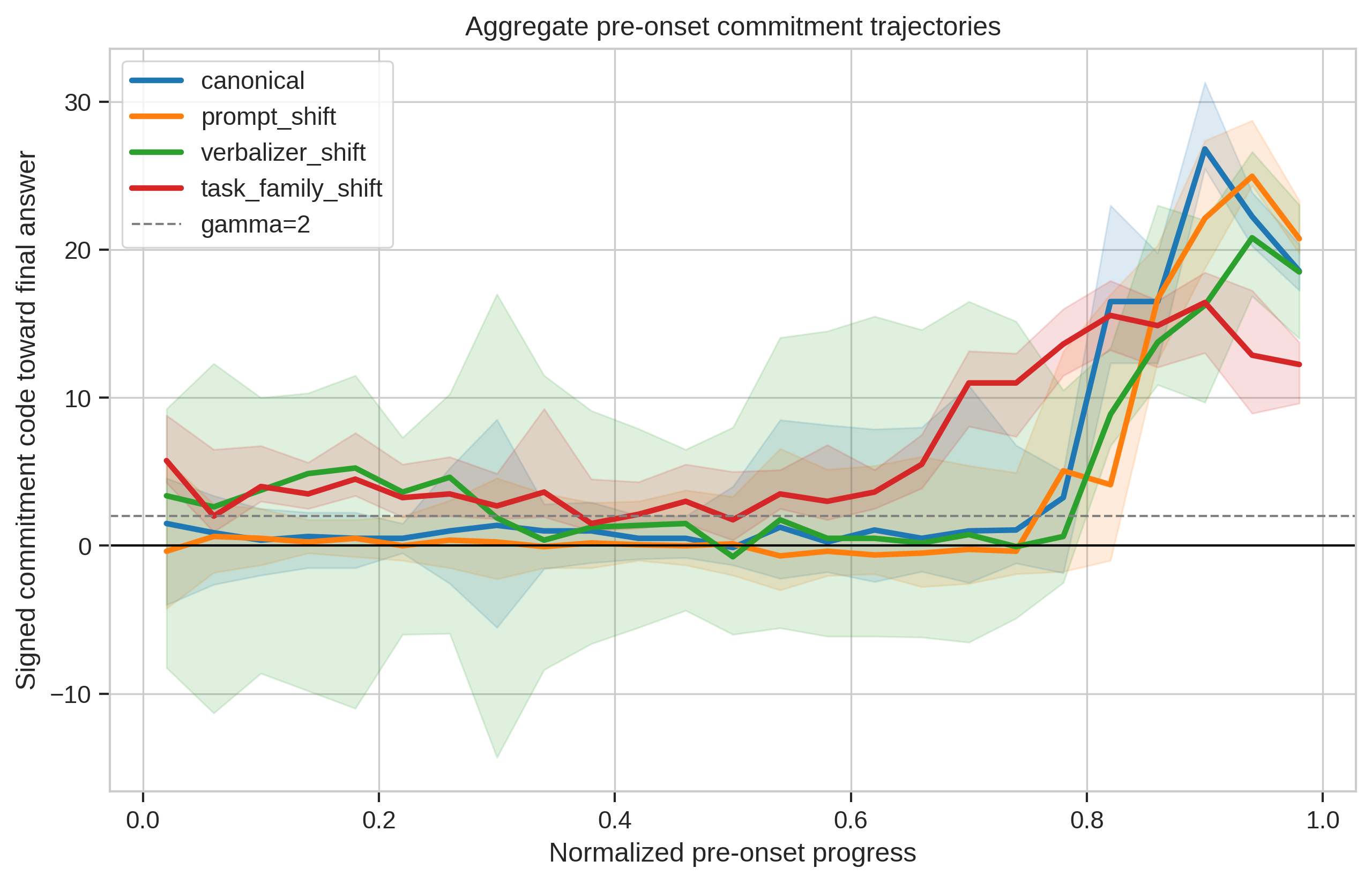}
\caption{\textbf{Aggregate signed-delta trajectories before answer onset.}
The y-axis is the commitment code signed so that positive values favor the eventual parsed answer. Lines show median trajectories over normalized pre-onset time, and bands show interquartile ranges. The horizontal dashed line is the main commitment threshold $\gamma=2$.}
\label{fig:app_signed_delta_atlas}
\end{figure}

\Cref{fig:app_signed_delta_atlas} shows that commitment is not only a scalar summary at one time point. Across conditions, the finite-answer projection tends to align with the eventual answer before the parser can observe the final verdict. This trajectory-level view also clarifies why lead can be large: in many samples the answer preference forms early, while the model continues to emit required intermediate template lines.

\subsection{Winner stability and margin growth}

\begin{figure}[h]
\centering
\includegraphics[width=\linewidth]{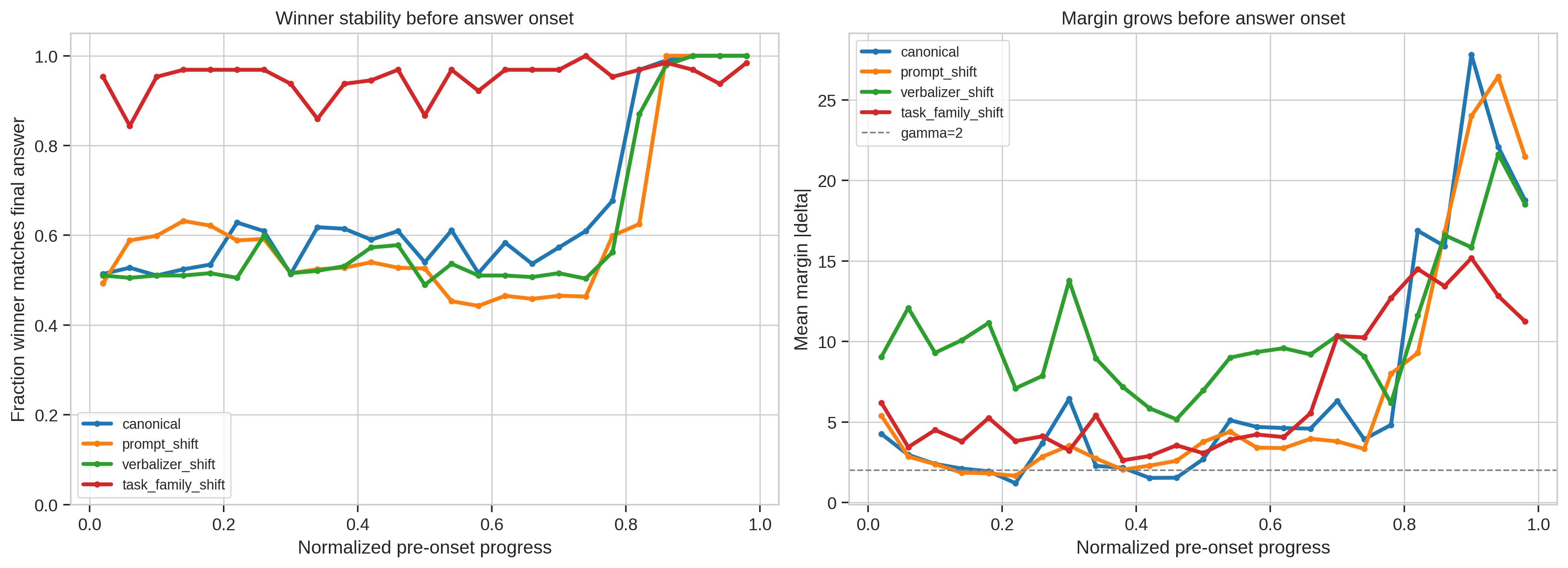}
\caption{\textbf{Winner stability and margin dynamics before answer onset.}
Left: fraction of states whose finite-answer winner matches the final parsed answer over normalized pre-onset time. Right: mean margin $|\delta_t|$ over the same interval. Commitment is not merely a transient sign crossing: winner agreement and margin are stable before the answer is verbalized.}
\label{fig:app_winner_stability_margin}
\end{figure}

The commitment-time definition requires that the winner not flip after commitment. \Cref{fig:app_winner_stability_margin} visualizes why this requirement is meaningful: winner agreement and margin both become stable before onset. This rules out the interpretation that lead is caused by isolated high-margin spikes followed by unstable preferences.

\section{Additional Diagnostics}
\label{app:additional_diagnostics}

This appendix collects diagnostic analyses that clarify what finite-answer commitment does and does not measure.

\subsection{Correctness and commitment}

\begin{table}[h]
\centering
\caption{\textbf{Correctness and retrospective stabilization are distinct.}
In task-family shift, wrong trajectories can still be retrospectively committed to the model's eventual wrong answer. Thus the measured signal tracks what the model is going to say, not whether that answer is true.}
\label{tab:app_correct_wrong}
\tabfit{
\begin{tabular}{llrrrrr}
\toprule
Condition & Group & $n$ & Mean lead & Median lead & Mean final margin & Mean sign flips \\
\midrule
Canonical & correct + committed & 96 & 17.41 & 10.0 & 20.73 & 4.26 \\
Prompt shift & correct + committed & 96 & 19.69 & 10.0 & 23.37 & 6.61 \\
Verbalizer shift & correct + committed & 96 & 17.19 & 9.0 & 19.30 & 2.41 \\
Task-family shift & correct + committed & 54 & 32.94 & 36.0 & 13.75 & 0.74 \\
Task-family shift & wrong + committed & 8 & 18.63 & 14.0 & 6.22 & 5.50 \\
Task-family shift & wrong + not committed & 2 & -- & -- & 0.88 & 9.00 \\
\bottomrule
\end{tabular}
}
\end{table}

\subsection{Cursor progress is not the whole story}

A possible concern is that the commitment code is merely a proxy for surface progress through the template. To visualize this concern, we plot signed commitment against normalized pre-onset progress.

\begin{figure}[h]
\centering
\includegraphics[width=\linewidth]{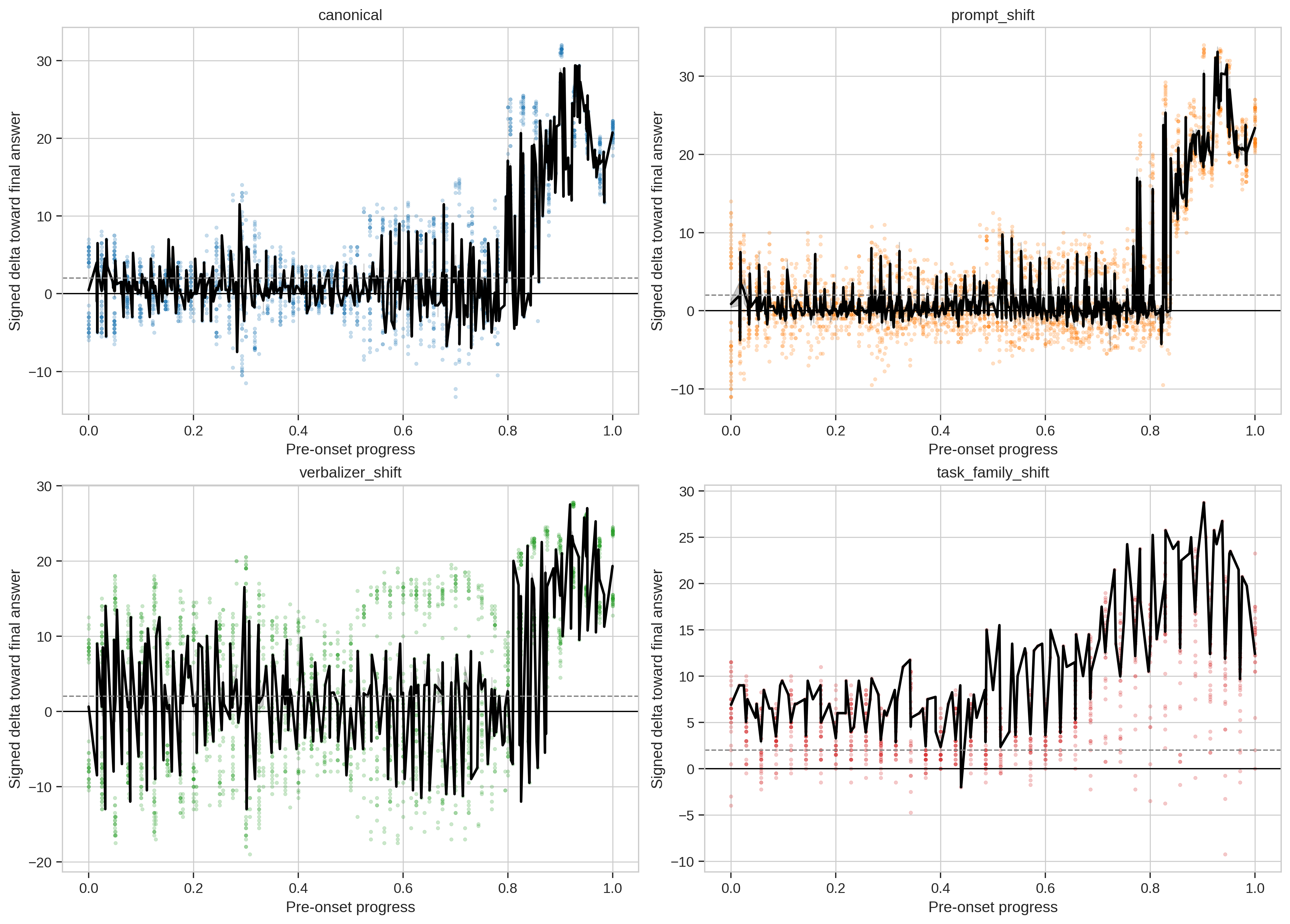}
\caption{\textbf{Commitment versus pre-onset cursor progress.}
Each panel plots signed commitment toward the final answer against normalized pre-onset progress. The black line shows the central trend. Although cursor progress and commitment are correlated in delayed-verdict templates, commitment varies substantially at fixed progress and is not reducible to a simple cursor variable.}
\label{fig:app_cursor_commit_diagnostic}
\end{figure}

\Cref{fig:app_cursor_commit_diagnostic} is not a replacement for the factorization analysis in the main paper. Instead, it provides an intuitive diagnostic for the cursor confound. The same cursor position can correspond to different commitment strengths, and commitment can be high before the final line begins. This motivates the operational factorization experiments, which more directly test whether commitment-dominant and cursor-dominant factors can be separated.

\subsection{Commitment to incorrect answers}

The commitment code measures the model's own projected answer preference, not ground truth. This distinction is visible in conditions where the model sometimes answers incorrectly.

\begin{figure}[h]
\centering
\includegraphics[width=\linewidth]{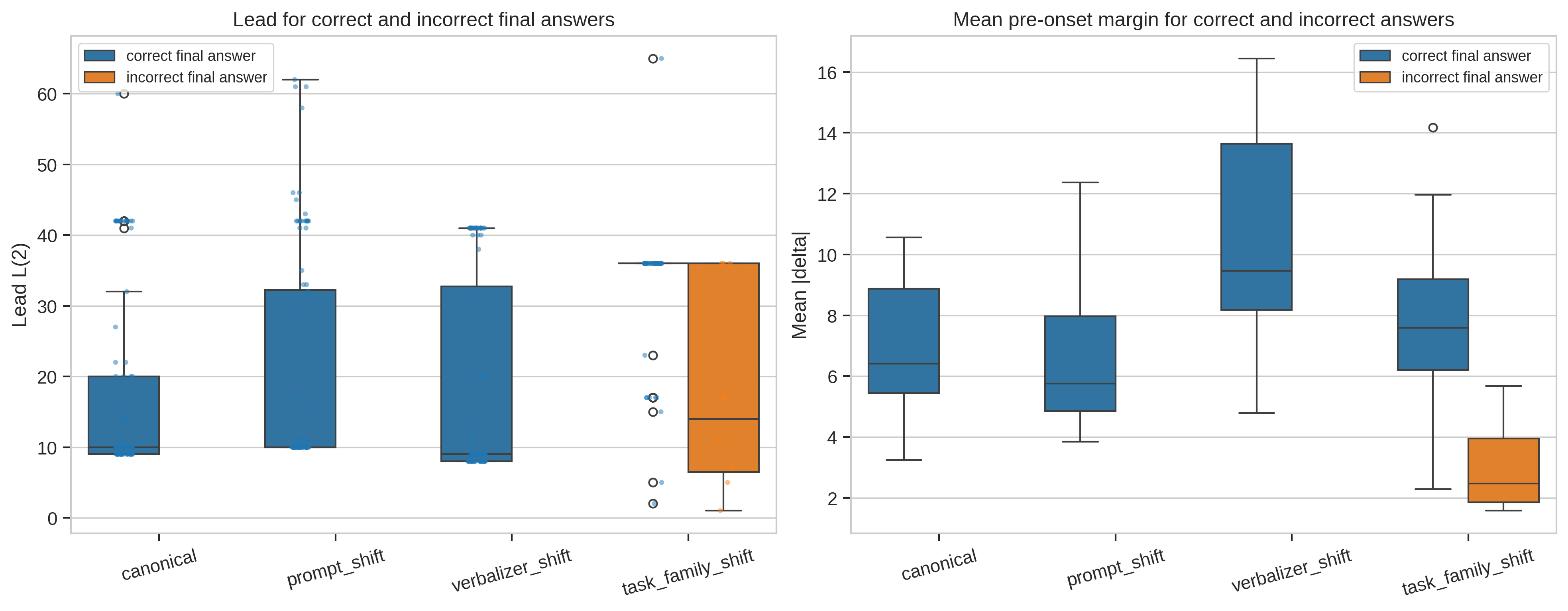}
\caption{\textbf{Commitment to correct and incorrect final answers.}
Left: lead distributions separated by whether the final parsed answer is correct. Right: mean pre-onset margin separated by correctness. Incorrect answers can also be committed before verbalization, confirming that finite-answer commitment is a model-internal preference measure rather than a truth detector.}
\label{fig:app_correct_vs_incorrect}
\end{figure}

\Cref{fig:app_correct_vs_incorrect} reinforces a central interpretive point. A model can commit early to a wrong answer. This is not a failure of the definition; it is precisely why commitment should not be identified with correctness. The finite-answer projection measures where the model's own continuation distribution is headed.

\section{Retrospective Stabilization Versus Online Detection}
\label{app:retrospective_online}

The main paper defines $\taucommit$ retrospectively: it is the earliest pre-onset state whose finite-answer winner matches the eventual parsed answer, exceeds margin threshold $\gamma$, and does not later flip before onset. This definition is appropriate for measuring when a trajectory first becomes stably aligned with its eventual answer, but it is not an online stopping rule.

To test this distinction, we evaluated a simple online rule that uses only current and past states: stop at the first position where $|\delta_t|\ge2$ and the last three available signs are identical. \Cref{tab:app_online_stop} shows that this rule stops in every trajectory, but often too early. In canonical, prompt-shift, and verbalizer-shift conditions, online stopping accuracy is close to chance because early high-margin signs can precede later flips. Task-family shift is different: it has fewer sign flips and the online rule is much more predictive.

\begin{table}[h]
\centering
\caption{\textbf{A naive online stopping rule is not equivalent to retrospective stabilization.}
The online rule stops at the first high-margin sign that is stable over the past three states. It has stop rate 1.0 in all conditions but low eventual-answer accuracy in parity-style conditions, showing that retrospective stabilization should not be interpreted as online detectability.}
\label{tab:app_online_stop}
\tabfit{
\begin{tabular}{lrrrrrrr}
\toprule
Condition & $n$ & Stop rate & Stop acc. & Mean online lead & Median online lead & Mean retro. lead & Mean sign flips \\
\midrule
Canonical & 96 & 1.000 & 0.521 & 42.92 & 42.0 & 17.41 & 4.26 \\
Prompt shift & 96 & 1.000 & 0.521 & 53.25 & 58.0 & 19.69 & 6.61 \\
Verbalizer shift & 96 & 1.000 & 0.510 & 41.20 & 41.0 & 17.19 & 2.41 \\
Task-family shift & 64 & 1.000 & 0.953 & 35.84 & 36.0 & 31.10 & 1.59 \\
\bottomrule
\end{tabular}
}
\end{table}

This analysis narrows the interpretation of the main result. The paper does not claim that an external observer can always declare commitment online from a prefix alone. Instead, it measures retrospective finite-answer preference stabilization along a completed pre-onset trajectory.

\section{Bare-Label Versus Contextual Verbalizer Scoring}
\label{app:bare_contextual}

The main experiments score condition-appropriate contextual verbalizers, such as final-answer-line continuations. A reviewer concern is that commitment measurements might depend strongly on the chosen verbalizers. To diagnose this, we rescored pre-onset states using bare-label verbalizers and compared them to the contextual $\delta$ used in the main analysis.

In the quality-gated parser-clean replication, bare-label and contextual $\delta$ values are highly correlated and usually agree in winner, but they are not identical. This supports a narrow interpretation: the measured object is a verbalizer-conditioned finite-answer projection, not a verbalizer-invariant semantic belief.

\begin{table}[h]
\centering
\caption{\textbf{Bare-label versus contextual scoring in the parser-clean replication.}
Bare-label and contextual finite-answer projections are highly correlated and usually agree in winner, but they are not identical. The measured object remains a verbalizer-conditioned finite-answer projection.}
\label{tab:app_v5_bare_contextual}
\tabfit{
\begin{tabular}{lrrrrr}
\toprule
Condition & $n$ & Corr. & Winner agree & Context final acc. & Bare final acc. \\
\midrule
Canonical & 250 & 0.998 & 1.000 & 0.668 & 0.668 \\
Prompt shift & 250 & 0.994 & 0.888 & 0.664 & 0.656 \\
Task-family shift & 192 & 1.000 & 0.995 & 0.719 & 0.714 \\
Verbalizer shift & 250 & 0.994 & 0.920 & 0.688 & 0.680 \\
\bottomrule
\end{tabular}
}
\end{table}

The result motivates our use of contextual verbalizer scoring. The finite-answer projection is defined relative to specified verbalizers, and changing those verbalizers can change the measured object. The main claim is therefore not that the model has a verbalizer-free semantic belief, but that its contextual finite-answer continuation preference can stabilize before the answer is parsed.

\section{Parser Diagnostics for the Main Conditions}
\label{app:parser_diag}

Because answer onset is parser-defined, parser failures could create selection bias. In the four main Qwen3 conditions, however, the strict parser succeeds on all samples. \Cref{tab:app_parser_diag} reports parse rates and output lengths. Thus the main results are not based on a parsed subset.

\begin{table}[h]
\centering
\caption{\textbf{Parser diagnostics for the main Qwen3 conditions.}
All main-condition outputs are parsed by the strict parser, so the main lead analysis does not suffer from parsed-subset selection bias.}
\label{tab:app_parser_diag}
\tabfit{
\begin{tabular}{lrrrr}
\toprule
Condition & Samples & Parse rate & Mean tokens & Mean chars \\
\midrule
Canonical & 96 & 1.000 & 43.09 & 87.52 \\
Prompt shift & 96 & 1.000 & 53.76 & 135.67 \\
Task-family shift & 64 & 1.000 & 36.55 & 111.11 \\
Verbalizer shift & 96 & 1.000 & 41.25 & 94.38 \\
\bottomrule
\end{tabular}
}
\end{table}
These diagnostics refer to the main delayed-verdict run reported in \Cref{tab:phenomenon}; the stricter quality-gated replication is reported separately in Appendix~\ref{app:quality_gated}.

\section{Quality-Gated Parser-Clean Replication}
\label{app:quality_gated}

Answer onset is parser-defined, so parser coverage is not a minor implementation detail. To test whether the main measurement survives without parsed-subset selection, we ran a stricter quality-gated replication with parser-compatible templates. The run proceeds to scoring only after passing a parser coverage gate.

\begin{table}[h]
\centering
\caption{\textbf{Quality-gated parser-clean replication.}
This stricter run achieves parse rate 1.0 in all conditions, removing parsed-subset selection bias. Retrospective finite-answer lead remains positive in every condition, although the measured lead is shorter than in the main delayed-verdict templates.}
\label{tab:app_v5_quality_gated}
\tabfit{
\begin{tabular}{lrrrrrrr}
\toprule
Condition & $n$ & Parse & Parsed acc. & Sample acc. & Commit & Mean onset & Mean lead [95\% CI] \\
\midrule
Canonical & 96 & 1.000 & 1.000 & 1.000 & 1.000 & 42.70 & 3.50 [3.19, 3.78] \\
Prompt shift & 96 & 1.000 & 1.000 & 1.000 & 1.000 & 55.99 & 4.74 [3.95, 5.65] \\
Task-family shift & 64 & 1.000 & 1.000 & 1.000 & 1.000 & 42.56 & 13.72 [10.78, 16.52] \\
Verbalizer shift & 96 & 1.000 & 1.000 & 1.000 & 1.000 & 52.01 & 4.22 [3.90, 4.55] \\
\bottomrule
\end{tabular}
}
\end{table}

\begin{figure}[h]
\centering
\includegraphics[width=0.82\linewidth]{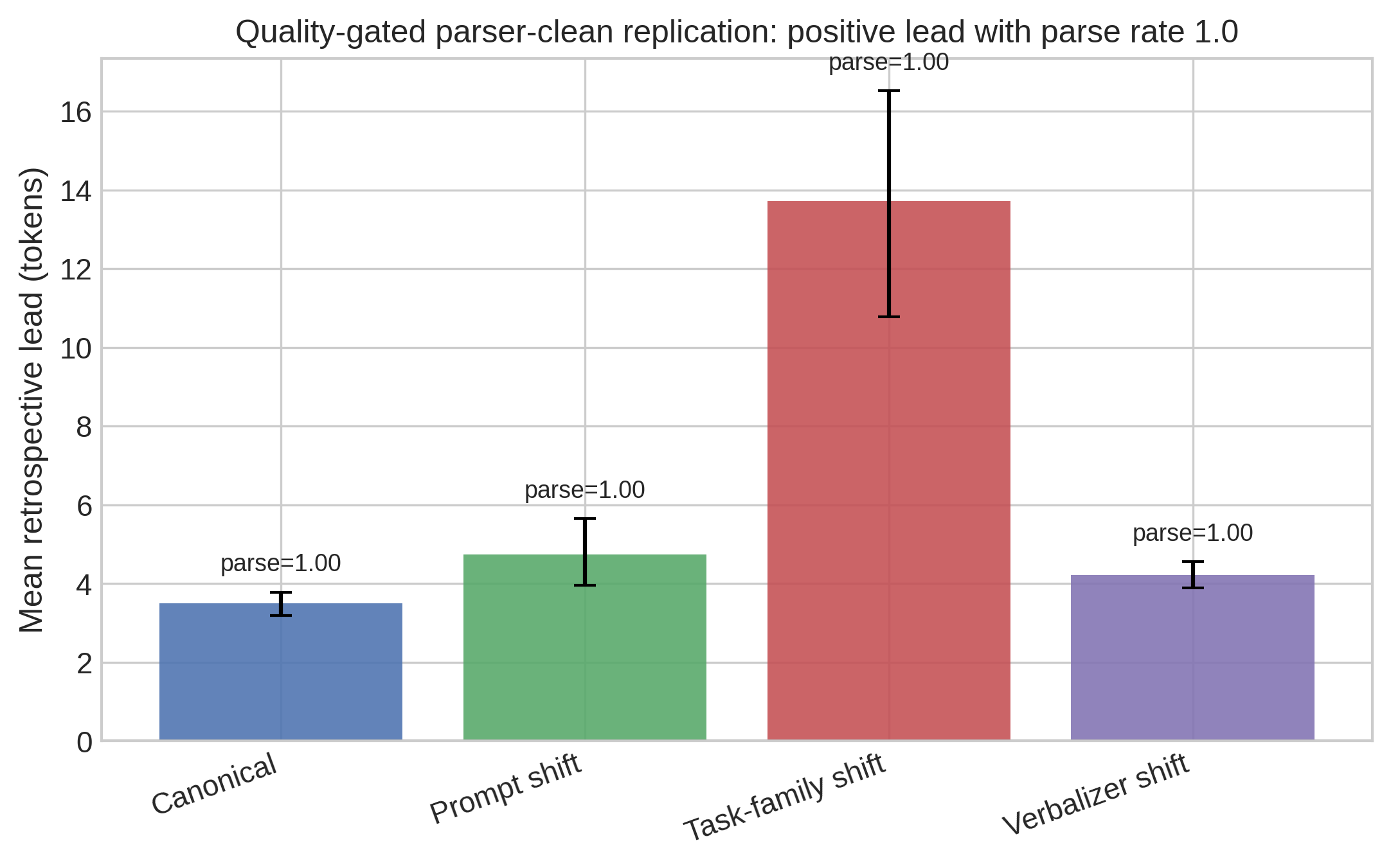}
\caption{\textbf{Quality-gated parser-clean replication.}
All conditions have parse rate 1.0 and positive retrospective lead. The lead is shorter than in the main delayed-verdict templates, showing that lead magnitude depends on template geometry while the existence of pre-onset stabilization does not depend on parsed-subset selection.}
\label{fig:app_v5_quality_gated}
\end{figure}

This replication changes the interpretation in a useful way. It does not aim to maximize lead length. Instead, it asks whether the measurement survives when output formatting is tightly controlled. The answer is yes: grouped-bootstrap lower bounds for mean lead are above zero in all conditions. At the same time, the shorter lead cautions against treating a specific number of tokens as a universal property of the model.

\begin{table}[h]
\centering
\caption{\textbf{Calibrated online detector in the parser-clean replication.}
Rules are selected on train trajectories and evaluated on held-out trajectories within each condition. This is an online prediction diagnostic, not the retrospective definition used for the main commitment time.}
\label{tab:app_v5_online}
\tabfit{
\begin{tabular}{lrrrrl}
\toprule
Condition & Test $n$ & Stop rate & Stop acc. & Mean lead & Selected rule \\
\midrule
Canonical & 29 & 1.000 & 1.000 & 6.62 & $\gamma=1$, progress $\ge 0.65$, window 2 \\
Prompt shift & 29 & 1.000 & 1.000 & 7.10 & $\gamma=5$, progress $\ge 0.65$, window 2 \\
Task-family shift & 19 & 1.000 & 0.895 & 18.05 & $\gamma=5$, progress $\ge 0.35$, window 1 \\
Verbalizer shift & 29 & 1.000 & 0.966 & 5.90 & $\gamma=8$, progress $\ge 0.75$, window 1 \\
\bottomrule
\end{tabular}
}
\end{table}

The online detector in \Cref{tab:app_v5_online} is a separate diagnostic. It shows that calibrated online rules can predict eventual parsed answers with high held-out accuracy in this controlled setting. We report stop rate, stop accuracy, and lead because each calibrated rule stops once per trajectory; precision/recall against retrospective commitment time is less informative here than whether the stopped answer matches the eventual parsed answer and how early it stops. This does not replace the retrospective definition; it shows that online detectability can be learned under calibrated rules.
The bare/contextual diagnostic is reported separately in Appendix~\ref{app:bare_contextual}. It shows high correlations and usually high winner agreement, while still confirming that the measured object is verbalizer-conditioned rather than a verbalizer-free semantic belief.

\section{Reproducibility Checklist}
\label{app:repro}

\begin{itemize}
\item Main model: \texttt{Qwen/Qwen3-4B-Instruct-2507}.
\item Decoding: greedy.
\item Quantization: main measurement and readout experiments use 4-bit NF4 inference; the exact residual-stream causal-sensitivity pilot uses full/bfloat16 inference.
\item Main conditions: canonical, prompt shift, verbalizer shift, task-family shift.
\item Main samples: 352 total.
\item Exact target: $\delta_t=S_\theta(\yes\mid\xii_t)-S_\theta(\no\mid\xii_t)$.
\item Hidden summaries: \texttt{last\_L21}, \texttt{concat\_selected}.
\item Readout: Ridge regression with grouped sample splits.
\item Factorizer: MLP encoder with 8-dimensional $u$ and $v$.
\item Controls: shuffled commitment labels and shuffled cursor labels.
\item Robustness: threshold sweep, multi-split readouts, sample-size scaling, layer sweep, additional prompt/task/verbalizer/free-form conditions, cross-model replications, and alignment-lite analysis.
\end{itemize}

\end{document}